\PassOptionsToPackage{table,dvipsnames}{xcolor}
\PassOptionsToPackage{sort&compress}{natbib}
\documentclass[11pt,a4paper]{thuc3i}
\usepackage{natbib}
\renewcommand{\today}{2026-11-22}
\setheadertext{JockeyBench}
\usepackage[T1]{fontenc}
\usepackage[utf8]{inputenc}
\usepackage[english]{babel}
\usepackage{microtype}
\usepackage{listings}
\usepackage[skins,breakable]{tcolorbox}
\usepackage{graphicx}
\usepackage{placeins}
\usepackage{booktabs}
\usepackage{amsmath}
\usepackage{amssymb}
\usepackage{pifont}
\usepackage{xcolor}
\usepackage{colortbl}
\usepackage{threeparttable}
\usepackage{multirow}
\usepackage{hyperref}
\usepackage{fontawesome5}
\newcommand{\CodePolicyWorkflow}{{\footnotesize\textcolor{RBBlue}{Code-as-policy}}}
\newcommand{\LLMPolicyWorkflow}{{\footnotesize\textcolor{RBOrange}{LLM-as-policy}}}

\usepackage{tabularx}
\usepackage{array}
\ifdefined\pdfglyphtounicode
  \pdfglyphtounicode{tfm:cmex10/parenleftBig}{0028 FE02}
  \pdfglyphtounicode{tfm:cmex10/parenrightBig}{0029 FE02}
  \pdfglyphtounicode{tfm:cmex10/parenleftbigg}{0028 FE03}
  \pdfglyphtounicode{tfm:cmex10/parenrightbigg}{0029 FE03}
  \pdfglyphtounicode{tfm:cmex10/bracketleftbigg}{005B FE03}
  \pdfglyphtounicode{tfm:cmex10/bracketrightbigg}{005D FE03}
  \pdfglyphtounicode{tfm:cmex10/summationtext}{2211 FE01}
  \pdfglyphtounicode{tfm:cmex10/summationdisplay}{2211 FE02}
  \pdfglyphtounicode{tfm:cmex10/hatwider}{02C6 FE02}
  \pdfglyphtounicode{tfm:cmex10/tildewide}{02DC FE01}
  \pdfglyphtounicode{tfm:cmex10/radicalBig}{221A FE02}
\fi
\definecolor{ComparisonTeal}{HTML}{14888D}
\definecolor{ComparisonMuted}{HTML}{68798C}

\newcommand{\nModels}{29}
\newcommand{\nARuns}{145}

\newcommand{\nRaces}{324}
\newcommand{\nDays}{10}

\newcommand{\bestModel}{GPT-6.1-Sol}

\newcommand{\allWinROI}{-21.3}

\newcommand{\nTakeoutRuns}{68}
\newcommand{\nTakeoutModels}{20}

\newcommand{\nHolmPos}{7}

\definecolor{RBBlue}{HTML}{466BA5}
\definecolor{RBOrange}{HTML}{C97840}
\definecolor{RBPurple}{HTML}{7352A0}
\definecolor{RBGreen}{HTML}{438267}
\definecolor{RBTeal}{HTML}{147D92}
\definecolor{RBMuted}{HTML}{4A5568}
\definecolor{RBRule}{HTML}{A6ADB7}
\newcommand{\rbcell}[3]{\cellcolor{#1!#2}#3}
\newcommand{\Live}{Reflect}
\newcommand{\Frozen}{Frozen}
\newcommand{\rbdashrule}{\noalign{\vskip 1.0pt\hbox to \textwidth{\leaders\hbox to 4pt{\vrule height \arrayrulewidth depth 0pt width 2pt\hss}\hfill}\vskip 1.6pt}}
\definecolor{RBHeatGreen}{HTML}{54AA4B}\definecolor{RBHeatPurple}{HTML}{942EB5}\definecolor{RBHeatBlue}{HTML}{2C4EC7}
\definecolor{RBHeatGreenTop}{HTML}{2E7A2A}\colorlet{RBHeatPurpleTop}{RBHeatPurple}\colorlet{RBHeatBlueTop}{RBHeatBlue}
\newcommand{\rbheat}[3]{\ifnum#2>99 \cellcolor{#1Top!\ifnum#2>100 80\else 70\fi}\color{white}\colorlet{RBMuted}{white!82!#1Top}\else\cellcolor{#1!#2}\fi #3}

\title{JockeyBench: The Racing Analyst's Last Exam}
\author{%
  Zhongwei Xie\textsuperscript{1,*}, Xingsu Li\textsuperscript{1,*},
  Jingqiao Yu\textsuperscript{1,*}, Zihao Wang\textsuperscript{2},
  Hong ChiTat\textsuperscript{1}, Zhaowei Wang\textsuperscript{1},
  Weizhi Du\textsuperscript{3}, Yuxuan Liu\textsuperscript{1},
  Huihao Jing\textsuperscript{1}, Jiaxin Bai\textsuperscript{4,\textdagger},
  Yangqiu Song\textsuperscript{1}
  \vskip4pt
  {\normalfont\fontsize{8.5}{11}\selectfont
    \textsuperscript{1}The Hong Kong University of Science and Technology\\
    \textsuperscript{2}Tsy Capital Limited\quad
    \textsuperscript{3}Independent researcher\quad
    \textsuperscript{4}Hong Kong Baptist University\par}
  \vskip3pt
  {\normalfont\fontsize{8.5}{11}\selectfont
    \texttt{\{zxiebk,xlijp,jyuck\}@connect.ust.hk}\\
    \texttt{zihaow0710@gmail.com, baijiaxin@hkbu.edu.hk, yqsong@ust.hk}\\
    \textsuperscript{*}Equal contribution.\quad
    \textsuperscript{\textdagger}Corresponding author.\par}
}
\resource{}{\href{https://github.com/HKUST-KnowComp/JockeyBench}{\faGithub\ \textbf{Code}}}

\begin{abstract}
Evaluating LLM agents in noisy sequential forecasting is challenging: strong performance may reflect a capable initial predictor rather than successful online revision, while accurate forecasts may still fail to deliver economic value after costs and market frictions. Existing benchmarks rarely isolate revision gains, compare reusable model-written predictors with direct LLM forecasting on the same tasks, or assess forecast quality alongside research costs and realized returns. To address these gaps, we introduce JockeyBench, a leakage-controlled replay benchmark built from 12 years of Japanese horse-racing records. Agents acquire evidence at a cost and seal probabilistic forecasts before outcomes are revealed. To isolate the effect of revision, we pair each adaptive \emph{code-as-policy} agent with a frozen copy of its initial predictor. We also compare code-as-policy, which executes reusable prediction programs, with \emph{LLM-as-policy}, which forecasts each race directly. We evaluate 29 models on 324 races. The strongest initial predictors outperform a conditional-logit baseline trained on 20,730 races. The two top-ranked models retain their initial predictors throughout evaluation. Revision helps weaker predictors correct overconfidence, but updates driven by noisy outcomes can hurt performance. \emph{Code-as-policy} generally produces more accurate forecasts at lower research cost, while models that struggle to write reliable predictors benefit from forecasting directly with \emph{LLM-as-policy}. Yet all 29 models, including Claude Opus 5.5 and GPT-6 Astra, have negative mean returns in retrospective settlement. JockeyBench thus offers a demanding testbed for current LLM agents.
\end{abstract}

\begin{document}
\maketitle

\section{Introduction}
\begin{figure}[t]
\centering
\includegraphics[width=\textwidth]{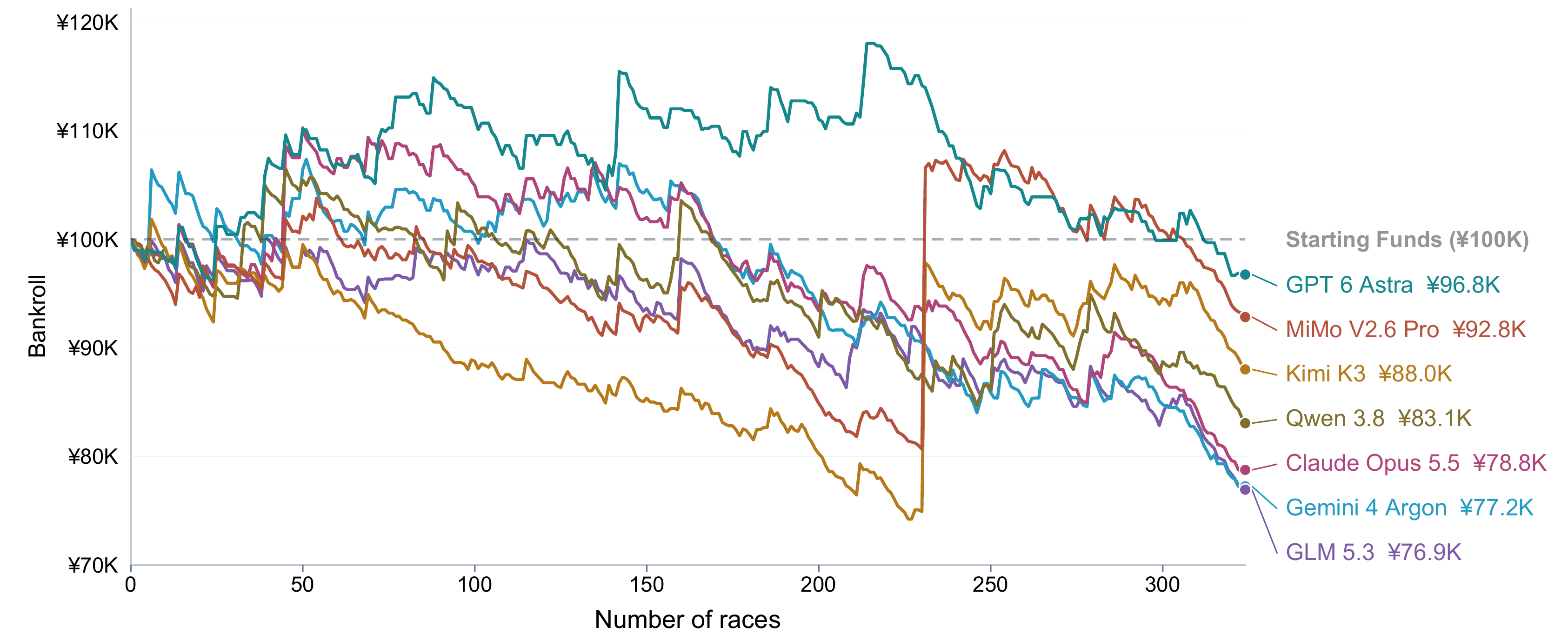}
\caption{\textbf{Model betting performance on \textsc{JockeyBench}.} \textsc{JockeyBench} tasks models with forecasting 324 Japanese horse races run on 10 race days in 2024 between January 6 and February 3. The figure shows the top code-as-policy configuration from each of the seven highest-ranked families by net utility (Table~\ref{tab:main}), tracking mean bankroll across 5 runs from JPY 100,000 under 1\%-capped Kelly WIN staking at official JRA dividends (Appendix~\ref{app:kelly}).}
\label{fig:teaser}
\end{figure}
Sequential forecasting of recurring events requires identifying predictive signals in noisy historical evidence. LLM agents must decide what evidence to acquire and how to construct and revise forecasting procedures as outcomes become available. An LLM agent can build a predictive model through code generation and experimentation~\citep{huang2023mlagentbench,xie2026coworkenv}, or forecast each event directly from evidence~\citep{halawi2024approaching,karger2025forecastbench}. 

Existing evaluations of such agents, however, leave three gaps. (1) \emph{Initial modeling is confounded with adaptation}: agents that revise as outcomes arrive are scored only on the trajectory they actually took, so a strong first predictor cannot be told apart from successful learning. (2) \emph{The place of LLM reasoning is fixed in advance}: model-building benchmarks let the LLM write code, forecasting benchmarks have it answer each question, and we know of none that compares the two on the same evidence. (3) \emph{Forecast quality, monetary value, and research cost are rarely measured together}: most forecasting benchmarks stop at probability scores and most trading benchmarks score returns without a proper score for the forecasts behind them. The value of a decision depends not only on forecast accuracy but also on the returns, risks, and costs of acting on the forecasts. Without these factors, we cannot reliably assess decision quality.

We introduce \textsc{JockeyBench} to close these gaps. Built from the Japan Racing Data Bank (JRDB) archive of Japanese horse racing~\citep{jrdbDataSpecifications}, it provides a leakage-controlled replay benchmark of 40,357 multi-runner races with linked runners, past performances, workouts, and results. The closing market supplies a strong reference forecast~\citep{benter1994computer}, and official dividends permit settlement under fixed betting rules. A restricted interface releases outcomes only after predictions are sealed~\citep{kapoor2023leakage}, and odds and expert ratings are withheld from agents. To address gap (1), we pair each revised code-as-policy trajectory with a frozen one, which keeps the same agent’s initial predictor for the whole forecasting task. To address gap (2), we compare two frameworks that use the same races, tools, and models but differ in how they generate each forecast. In code-as-policy, the agent writes and refines a program that predicts the outcome of each race. In LLM-as-policy, the agent maintains an evidence plan that selects the inputs for each race, and the model reads those inputs and forecasts every race by a single call. To address gap (3), we evaluate forecast quality, research cost, and monetary value separately. We score forecasts using a logarithmic scoring rule, charge agents for the evidence they acquire so that research costs enter their objective, and settle bets derived from the same forecasts at official dividends. Across \nARuns{} paired runs of \nModels{} configurations and a nine-model workflow comparison on \nRaces{} races, our experiments trace a single narrative:
\begin{itemize}
\setlength{\leftmargin}{0em}
\setlength{\itemsep}{1pt}
\setlength{\parskip}{0pt}
\setlength{\topsep}{2pt}
\item \textbf{Initial feature design and calibration drive predictor quality.} The strongest initial predictors beat a 20,730-race logit baseline and hold up under schema blinding, and an audit of 431 code-as-policy \Live{} trajectories finds no evidence of result recall (\S\ref{sec:quality}).
\item \textbf{Revision repairs weak predictors but rarely improves strong ones.} No Tier-1 configuration changes significantly; below it, \nHolmPos{} configurations gain by fixing overconfidence and 3 decline by overreacting to biased warmup samples or daily noise (\S\ref{sec:revision}).
\item \textbf{Code-as-policy delivers higher forecast quality at lower research cost, while LLM-as-policy helps when model-written predictors are unreliable.} Across nine shared models, code-as-policy leads on log gain for seven and incurs lower research fees for all nine (\S\ref{sec:code-vs-llm-policy}).
\end{itemize}

Yet, as Figure~\ref{fig:teaser} shows, none of these improvements in forecasting performance translate into profitability, even for frontier models such as Claude Opus 5.5 and GPT-6 Astra. Detailed analyses are in \S\ref{sec:workflow-betting}.
Every configuration loses money on average under capped Kelly at the JRA's 20\% takeout. Across all nine models in the matched comparison, both workflows overestimate the win rate of their Kelly selections.

\section{Related Work}
\label{sec:related}
\begin{table}[t]
\caption{\textbf{Comparison of \textsc{JockeyBench} with related benchmarks.}
\emph{Code}: agent-written predictor code reused across decisions;
\emph{Strat.\ Rev.}: agent explicitly revises its strategy after outcomes are released;
\emph{Score}: scores probability forecasts with a proper scoring rule, such as the Brier, log, or interval score;
\emph{Replay}: same events can be re-run for new models;
\emph{Mem.\ Ctrl.}: guards inputs against pretraining outcome recall by using events after model cutoffs or de-identified records; instructions alone do not count.}
\label{tab:related}
\centering
\fontsize{7.2}{8.4}\selectfont
\setlength{\tabcolsep}{2.3pt}
\renewcommand{\arraystretch}{0.95}
\newcommand{\rwy}{\textcolor[HTML]{64CE46}{\ding{51}}}
\newcommand{\rwn}{\textcolor[HTML]{CE4E58}{\ding{55}}}
\begin{tabular*}{\textwidth}{@{\extracolsep{\fill}}lcccccccc@{}}
\toprule
\textbf{Benchmark} & \textbf{Scale} & \textbf{Models} & \textbf{Runs} & \textbf{Code} & \textbf{Strat.\ Rev.} & \textbf{Score} & \textbf{Replay} & \textbf{Mem.\ Ctrl.} \\
\midrule
\multicolumn{9}{@{}l}{\textit{Prediction-market forecasting}} \\
Prophet Arena~\citep{yang2025prophet} & 1,367 events & 23 & 1 & \rwn & \rwn & \rwy & \rwn & \rwy \\
PolyBench~\citep{cheng2026polybench} & 5,097 markets & 7 & 1 & \rwn & \rwn & \rwn & \rwy & \rwy \\
\midrule
\multicolumn{9}{@{}l}{\textit{Financial trading and portfolio evaluation}} \\
InvestorBench~\citep{li2025investorbench} & 10 assets & 13 & 5 & \rwn & \rwy & \rwn & \rwy & \rwn \\
DeepFund~\citep{li2025deepfund} & 5 stocks, 24 days & 9 & 1 & \rwn & \rwn & \rwn & \rwn & \rwy \\
FinSABER~\citep{li2026finsaber} & 63--91 stocks & 4 & 1 & \rwn & \rwy & \rwn & \rwy & \rwn \\
Investor's Last Exam~\citep{jiang2026ile} & 86 assignments & 11 & 1 & \rwn & \rwn & \rwy & \rwy & \rwn \\
\midrule
\multicolumn{9}{@{}l}{\textit{Sequential sports betting}} \\
KellyBench~\citep{grady2026kellybench} & 120 matchdays & 5 & 5 & \rwy & \rwy & \rwy & \rwy & \rwn \\
WC2026-Agents~\citep{ding2026wc2026} & 104 matches & 4 & 1 & \rwn & \rwy & \rwy & \rwn & \rwy \\
\midrule
\textbf{\textsc{JockeyBench} (Ours)} & 324 races & \textbf{29} & \textbf{5} & \rwy & \rwy & \rwy & \rwy & \rwy \\
\bottomrule
\end{tabular*}
\end{table}

\begin{figure}[t]
\centering
\includegraphics[width=\textwidth]{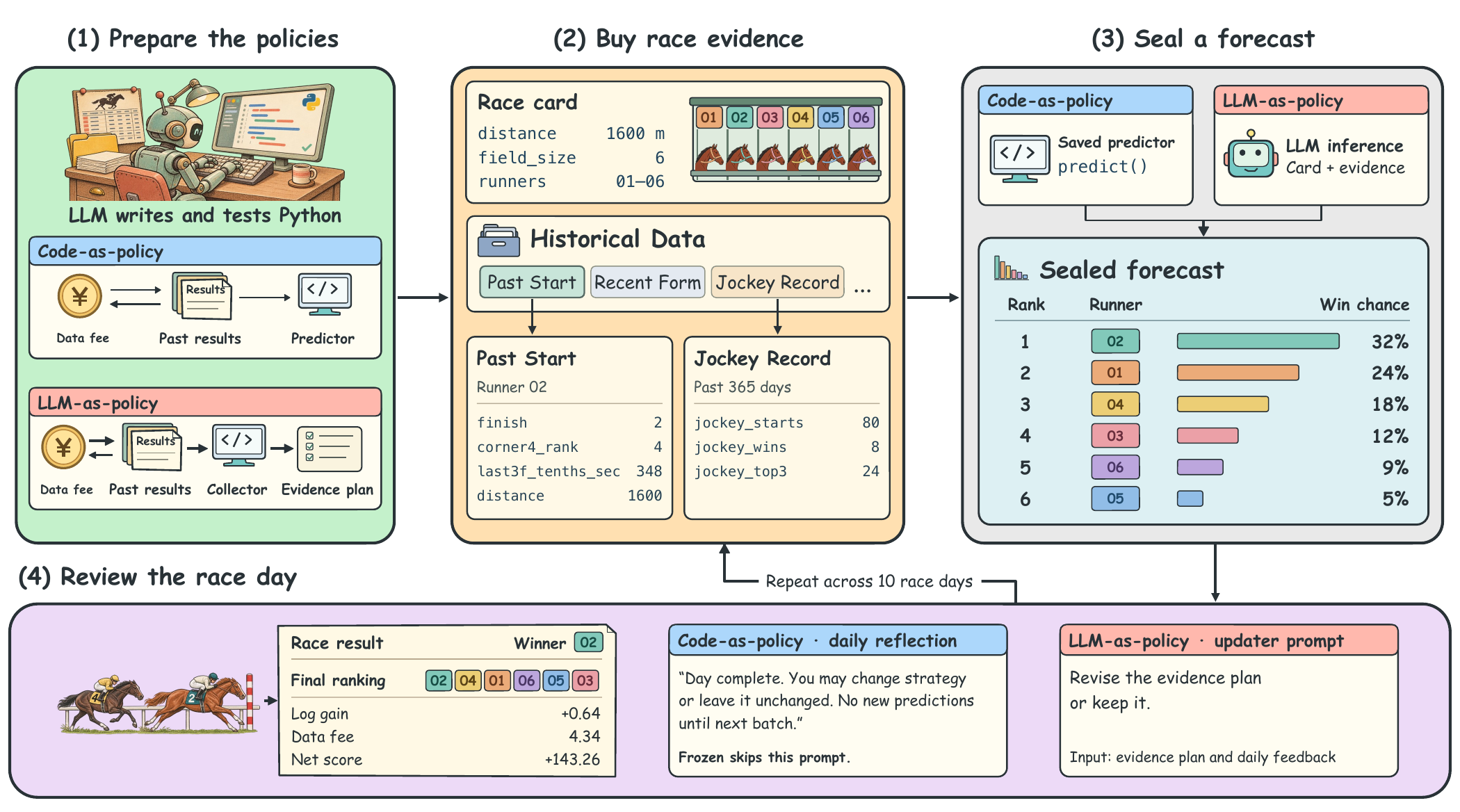}
\caption{\textbf{The \textsc{JockeyBench} pipeline.} (1)~Before the evaluation window, the agent writes and tests Python on past results, paying for the evidence it reads. On each race day, it (2)~buys evidence for the day's races and (3)~seals a ranked forecast before the results are released: code-as-policy runs its saved \texttt{predict()}, and LLM-as-policy passes the race card and collected evidence to one LLM call. (4)~After release, the agent may revise its predictor or evidence plan; the frozen control skips this step. Steps 2--4 repeat over 10 race days.}
\label{fig:framework}
\end{figure}

Table~\ref{tab:related} compares \textsc{JockeyBench} with related betting and trading evaluations. Each feature appears in prior work; our difference is their combination.

\paragraph{Executable code, revision, and cost-aware agents.}
Agents build machine-learning solutions in MLAgentBench, MLE-bench, and AIDE~\citep{huang2023mlagentbench,chan2024mle,jiang2025aide} and sustain long-horizon business decisions in Vending-Bench~\citep{backlund2025vending}. Building on code-as-policies~\citep{liang2023code} and cognitive views of code as procedural memory~\citep{sumers2024cognitive}, recent agents store executable skills and world models~\citep{xiesurvey2026, wang2023voyager,wang2025inducing,tang2024worldcoder} or evolve and self-modify code against fixed evaluators~\citep{romeraparedes2024mathematical,hu2024automated,yin2025godel,zhang2026darwin}. By contrast, methods such as Reflexion and ACE revise textual memory rather than executable programs~\citep{shinn2023reflexion,zhang2025ace, huang2026towards}. Cost-aware evaluation prices tools and reasoning~\citep{kapoor2024agents,liu2025costbench, xie2026gala}. Unlike offline code search against fixed evaluators, \textsc{JockeyBench} revises deployed predictors under noisy, time-ordered outcomes, prices every evidence query, and isolates revision with a frozen copy of the first predictor.

\paragraph{Forecasting and financial decision-making.}
LLM forecasting benchmarks score probabilities on unresolved questions and live events~\citep{karger2025forecastbench,yang2025prophet,gao2026toward}, and PolyBench scores directional accuracy and confidence-weighted returns on prediction markets~\citep{cheng2026polybench}. Trading benchmarks score returns from memory-based agents~\citep{li2025investorbench,li2025deepfund,li2026finsaber,jiang2026ile}, and quantitative R\&D agents write factor code offline~\citep{tang2025alphaagent,li2025rdquant}. In sports betting, KellyBench lets agents revise betting models over a season~\citep{grady2026kellybench}, and WC2026-Agents runs web-enabled agents on live World Cup matches~\citep{ding2026wc2026}. Closest to our workflow comparison, \citet{tan2024language} find that removing LLM components from time-series forecasters need not reduce accuracy.

\paragraph{Horse-race modeling and market efficiency.}
Conditional-logit models of race winners~\citep{bolton1986searching,chapman1994still} and combinations of fundamental and market probabilities~\citep{benter1994computer} motivate our fitted baselines. Racetrack win pools have long exhibited a favorite--longshot bias~\citep{ali1977probability,thaler1988anomalies}; separately, beating a uniform prior does not imply beating the market price, so a forecaster can substantially outperform uniform and still lose money after takeout, as Section~\ref{sec:workflow-betting} shows.

\section{The JockeyBench Protocol}
\label{sec:env}
Figure~\ref{fig:framework} summarizes the \textsc{JockeyBench} protocol. Agents forecast a chronological sequence of races through a leakage-controlled replay environment (\S\ref{sec:data}), operating under either code-as-policy (\S\ref{sec:framework}) or LLM-as-policy (\S\ref{sec:llm-as-policy}). Their sealed predictions are then evaluated jointly on forecast quality, research cost, and retrospective betting returns (\S\ref{sec:score}).

\subsection{Replay Environment and Information Controls}
\label{sec:data}

\begin{figure}[!t]
\centering
\includegraphics[width=\textwidth]{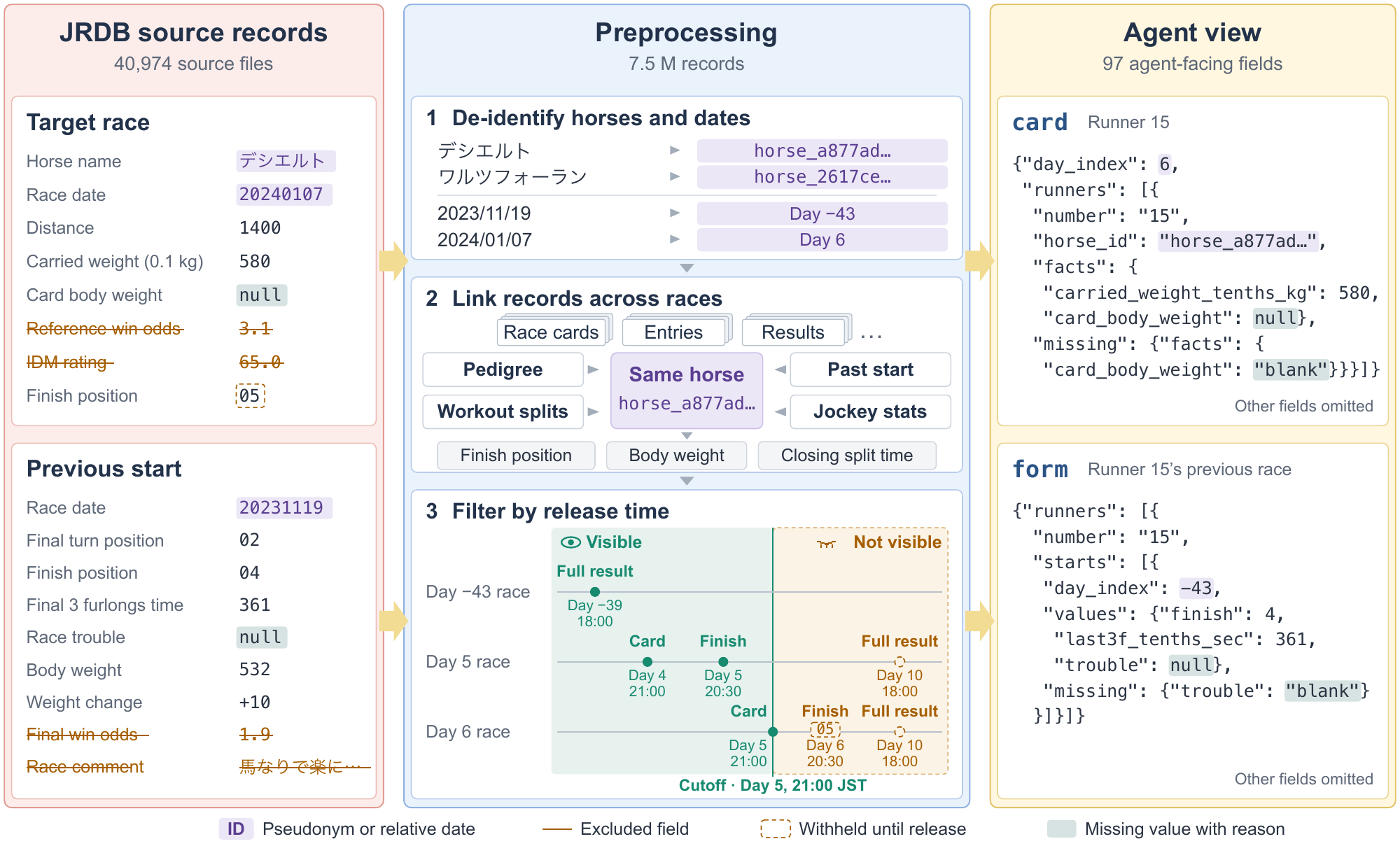}
\caption{\textbf{Preprocessing and release-time filtering for one runner and its previous start.} (1) Horse names and calendar dates become pseudonymous identifiers and relative day indices. (2) Records from different source files are linked through the shared horse identifier. (3) Each field becomes visible only after its scheduled release. Struck-through fields are excluded, and the dashed result is withheld until release. The \texttt{card} and \texttt{form} views show relative dates, pseudonymous identifiers, typed values, and explicit missing-value reasons; other fields are omitted.}
\label{fig:data-cleaning}
\end{figure}

\paragraph{Leakage-controlled replay archive.}
Evaluating agents on historical events requires strict control over both information timing and persistent state to prevent hindsight leakage~\citep{kapoor2023leakage,zhou2026historical}. We built \textsc{JockeyBench} by parsing, cleaning, and linking 12 years of raw JRDB Japanese horse racing files from 2015 to 2026, covering 40,357 races and 564,224 runner entries with their horses, jockeys, trainers, and pedigrees, into a time-gated replay database. Every record is stamped with its historical public release time, entity identities are HMAC-pseudonymized, and external network or tool access is disabled. To ensure forecasts rely on factual handicapping rather than market consensus, the agent-facing schema exposes 97 whitelisted objective fields, while withholding all market odds, popularity, dividends, expert ratings, and free-text notes. We withhold odds deliberately. When a matched arm shows the pre-race odds, every tested model converges on the market price, and the benchmark no longer separates them (Appendix~\ref{app:odds}). Figure~\ref{fig:data-cleaning} traces one runner from the raw records to the race card and past form available at the prediction cutoff. Appendix Table~\ref{tab:fields} lists the exposed fields.

\subsection{Framework 1: Code-as-policy}
\label{sec:framework}

In the first framework, reasoning of the LLM is compiled into executable code, extending the code-as-policies paradigm~\citep{liang2023code} to sequential forecasting. Following the prequential principle~\citep{dawid1984present,vzliobaite2015evaluation} and agentic code experimentation~\citep{huang2023mlagentbench,jiang2025aide}, the agent writes and maintains an executable Python predictor \texttt{predict(sdk, context)}, which alone produces every forecast. An exploration session first inspects labeled historical races and saves an initial predictor $P_0$. Each race day $d \in \{1,\dots,\nDays\}$ then runs one cycle of the loop in Figure~\ref{fig:framework}. When the race cards of the day become visible, the saved predictor seals a \textbf{ranking} and \textbf{win probabilities} for every race before post time. The races are then run, and their results are revealed and scored. Finally, a reflection session shows the agent the scores and costs of that day, and the agent either keeps its predictor or revises it. Appendices~\ref{app:workflows}, \ref{app:allowances}, and~\ref{app:validation} detail session timing, budgets, and predictor validation and fallback. Each \Live{} trajectory is paired with a \Frozen{} control that keeps the same initial predictor $P_0$ fixed across all \nRaces{} races.

\subsection{Framework 2: LLM-as-policy}
\label{sec:llm-as-policy}

In the second framework, the LLM decides each race itself. Whereas \emph{code-as-policy} persists a statistical predictor, \emph{LLM-as-policy} persists an executable evidence plan $P_d$, the function \texttt{collect\_evidence(sdk, context)}, and delegates probability estimation to a fresh per-race LLM call, as Figure~\ref{fig:framework} shows. Warmup before the window produces the initial plan $P_0$; before each race on day $d$, the current plan $P_{d-1}$ executes without an LLM call. Observing metadata on feature coverage, missingness, freshness, and query costs, $P_{d-1}$ queries track context once and selectively purchases recent history, workouts, and condition statistics. A stateless LLM session then receives the basic race card and acquired evidence package and returns a sealed ranking and win probabilities, with at most one repair request.

Once results are released, a daily update session reviews the \emph{Experience Frame}, a structured summary of daily and cumulative acquisition counts, fees, and forecast scores, and either retains $P_{d-1}$ or validates an updated evidence plan $P_d$ for day $d+1$. Appendices~\ref{app:active-research} and~\ref{app:acquisition-tariff} detail the update session and its tariff.

\begin{table}[t]
\caption{\textbf{Main results.} Each row averages five runs, ranked by native net utility. In the \Live{}$-$\Frozen{} column, teal marks significant gains and orange marks significant declines after Holm correction. In the shaded metric columns, white text marks the two highest model values and bold marks the highest.}
\label{tab:main}
\centering
\begingroup
\fontsize{7.2}{8.7}\selectfont
\setlength{\tabcolsep}{3.0pt}
\renewcommand{\arraystretch}{0.92}
\begin{tabular*}{\textwidth}{@{\extracolsep{\fill}}lcccccccc@{}}
\toprule
& \textcolor{RBHeatGreen}{\textbf{Utility}} & \multicolumn{3}{c}{\textcolor{RBHeatPurple}{\textbf{Forecast quality}}} & \multicolumn{2}{c}{\textcolor{RBTeal}{\textbf{Daily revision}}} & \multicolumn{2}{c}{\textcolor{RBHeatBlue}{\textbf{Market return}}} \\
\cmidrule(lr){2-2}\cmidrule(lr){3-5}\cmidrule(lr){6-7}\cmidrule(l){8-9}
Model / reference & \shortstack[c]{Net $\uparrow$\\/race} & \shortstack[c]{Log gain $\uparrow$\\mean $\pm$ SD} & \shortstack[c]{Top-1\\\% $\uparrow$} & \shortstack[c]{Top-3\\\% $\uparrow$} & \shortstack[c]{\Live{}$-$\Frozen{}\\mean $\pm$ SE} & \shortstack[c]{Predictors\\used} & \shortstack[c]{Kelly\\\% $\uparrow$} & \shortstack[c]{Flat WIN\\\% $\uparrow$} \\
\midrule
GPT-6-Astra & \rbheat{RBHeatGreen}{101}{$\mathbf{+61.3}$} & \rbheat{RBHeatPurple}{100}{$\phantom{+}{0.523}$ {\fontsize{5.8}{6.5}\selectfont\textcolor{RBMuted}{$\pm 0.008$}}} & \rbheat{RBHeatPurple}{101}{$\mathbf{30.0}$} & \rbheat{RBHeatPurple}{100}{${48.0}$} & $+0.000$ {\fontsize{5.8}{6.5}\selectfont\textcolor{RBMuted}{$\pm 0.000$}} & 1.0 & \rbheat{RBHeatBlue}{101}{$\mathbf{-3.2}\phantom{0}$} & \rbheat{RBHeatBlue}{100}{${-7.8}\phantom{0}$} \\
GPT-6.1-Sol & \rbheat{RBHeatGreen}{100}{${+59.5}$} & \rbheat{RBHeatPurple}{101}{$\phantom{+}\mathbf{0.525}$ {\fontsize{5.8}{6.5}\selectfont\textcolor{RBMuted}{$\pm 0.002$}}} & \rbheat{RBHeatPurple}{100}{${28.5}$} & \rbheat{RBHeatPurple}{101}{$\mathbf{48.3}$} & $+0.000$ {\fontsize{5.8}{6.5}\selectfont\textcolor{RBMuted}{$\pm 0.000$}} & 1.0 & \rbheat{RBHeatBlue}{37}{${-12.4}$} & \rbheat{RBHeatBlue}{39}{${-10.9}$} \\
Gemini 4 Argon & \rbheat{RBHeatGreen}{43}{${+54.2}$} & \rbheat{RBHeatPurple}{44}{$\phantom{+}{0.471}$ {\fontsize{5.8}{6.5}\selectfont\textcolor{RBMuted}{$\pm 0.017$}}} & \rbheat{RBHeatPurple}{40}{${27.4}$} & \rbheat{RBHeatPurple}{41}{${45.9}$} & $+0.001$ {\fontsize{5.8}{6.5}\selectfont\textcolor{RBMuted}{$\pm 0.003$}} & 1.8 & \rbheat{RBHeatBlue}{25}{${-22.8}$} & \rbheat{RBHeatBlue}{26}{${-16.9}$} \\
Claude Opus 5.5 & \rbheat{RBHeatGreen}{42}{${+53.1}$} & \rbheat{RBHeatPurple}{42}{$\phantom{+}{0.453}$ {\fontsize{5.8}{6.5}\selectfont\textcolor{RBMuted}{$\pm 0.030$}}} & \rbheat{RBHeatPurple}{40}{${27.5}$} & \rbheat{RBHeatPurple}{42}{${46.0}$} & $-0.001$ {\fontsize{5.8}{6.5}\selectfont\textcolor{RBMuted}{$\pm 0.001$}} & 1.2 & \rbheat{RBHeatBlue}{27}{${-21.2}$} & \rbheat{RBHeatBlue}{34}{${-13.1}$} \\
GPT-6.1-Terra & \rbheat{RBHeatGreen}{42}{${+52.6}$} & \rbheat{RBHeatPurple}{43}{$\phantom{+}{0.460}$ {\fontsize{5.8}{6.5}\selectfont\textcolor{RBMuted}{$\pm 0.028$}}} & \rbheat{RBHeatPurple}{38}{${26.9}$} & \rbheat{RBHeatPurple}{43}{${46.3}$} & $+0.006$ {\fontsize{5.8}{6.5}\selectfont\textcolor{RBMuted}{$\pm 0.005$}} & 2.0 & \rbheat{RBHeatBlue}{25}{${-22.6}$} & \rbheat{RBHeatBlue}{27}{${-16.5}$} \\
Claude Opus 5 & \rbheat{RBHeatGreen}{40}{${+51.4}$} & \rbheat{RBHeatPurple}{44}{$\phantom{+}{0.469}$ {\fontsize{5.8}{6.5}\selectfont\textcolor{RBMuted}{$\pm 0.039$}}} & \rbheat{RBHeatPurple}{40}{${27.5}$} & \rbheat{RBHeatPurple}{40}{${45.7}$} & $-0.001$ {\fontsize{5.8}{6.5}\selectfont\textcolor{RBMuted}{$\pm 0.003$}} & 3.4 & \rbheat{RBHeatBlue}{34}{${-14.9}$} & \rbheat{RBHeatBlue}{35}{${-12.8}$} \\
Claude Fable 5.1 & \rbheat{RBHeatGreen}{40}{${+50.7}$} & \rbheat{RBHeatPurple}{42}{$\phantom{+}{0.456}$ {\fontsize{5.8}{6.5}\selectfont\textcolor{RBMuted}{$\pm 0.048$}}} & \rbheat{RBHeatPurple}{37}{${26.7}$} & \rbheat{RBHeatPurple}{40}{${45.6}$} & $+0.007$ {\fontsize{5.8}{6.5}\selectfont\textcolor{RBMuted}{$\pm 0.011$}} & 5.0 & \rbheat{RBHeatBlue}{22}{${-26.2}$} & \rbheat{RBHeatBlue}{27}{${-16.3}$} \\
GPT-5.6-Sol & \rbheat{RBHeatGreen}{40}{${+50.6}$} & \rbheat{RBHeatPurple}{44}{$\phantom{+}{0.473}$ {\fontsize{5.8}{6.5}\selectfont\textcolor{RBMuted}{$\pm 0.023$}}} & \rbheat{RBHeatPurple}{41}{${27.8}$} & \rbheat{RBHeatPurple}{41}{${45.9}$} & $+0.004$ {\fontsize{5.8}{6.5}\selectfont\textcolor{RBMuted}{$\pm 0.004$}} & 4.2 & \rbheat{RBHeatBlue}{38}{${-11.7}$} & \rbheat{RBHeatBlue}{42}{${-9.8}\phantom{0}$} \\
GPT-6-Sol & \rbheat{RBHeatGreen}{39}{${+50.2}$} & \rbheat{RBHeatPurple}{44}{$\phantom{+}{0.471}$ {\fontsize{5.8}{6.5}\selectfont\textcolor{RBMuted}{$\pm 0.019$}}} & \rbheat{RBHeatPurple}{39}{${27.3}$} & \rbheat{RBHeatPurple}{39}{${45.4}$} & $+0.009$ {\fontsize{5.8}{6.5}\selectfont\textcolor{RBMuted}{$\pm 0.004$}} & 4.2 & \rbheat{RBHeatBlue}{36}{${-13.4}$} & \rbheat{RBHeatBlue}{40}{${-10.8}$} \\
\rbdashrule
\rowcolor{black!4}
\textit{Large-sample logit baseline} & -- & $\phantom{+}{0.462}$\phantom{ {\fontsize{5.8}{6.5}\selectfont\textcolor{RBMuted}{$\pm 0.000$}}} & ${25.3}$ & ${45.9}$ & -- & -- & ${-32.8}$ & ${-25.0}$ \\
\rbdashrule
GLM-5.3 & \rbheat{RBHeatGreen}{30}{${+38.6}$} & \rbheat{RBHeatPurple}{36}{$\phantom{+}{0.401}$ {\fontsize{5.8}{6.5}\selectfont\textcolor{RBMuted}{$\pm 0.054$}}} & \rbheat{RBHeatPurple}{30}{${24.6}$} & \rbheat{RBHeatPurple}{33}{${43.5}$} & $+0.017$ {\fontsize{5.8}{6.5}\selectfont\textcolor{RBMuted}{$\pm 0.010$}} & 5.8 & \rbheat{RBHeatBlue}{25}{${-23.1}$} & \rbheat{RBHeatBlue}{27}{${-16.2}$} \\
Qwen3.8-2.4T-A95B & \rbheat{RBHeatGreen}{29}{${+37.7}$} & \rbheat{RBHeatPurple}{33}{$\phantom{+}{0.367}$ {\fontsize{5.8}{6.5}\selectfont\textcolor{RBMuted}{$\pm 0.048$}}} & \rbheat{RBHeatPurple}{30}{${24.6}$} & \rbheat{RBHeatPurple}{32}{${43.1}$} & $+0.030$ {\fontsize{5.8}{6.5}\selectfont\textcolor{RBMuted}{$\pm 0.012$}} & 4.2 & \rbheat{RBHeatBlue}{32}{${-16.9}$} & \rbheat{RBHeatBlue}{34}{${-13.3}$} \\
MiMo-V2.6-Pro & \rbheat{RBHeatGreen}{27}{${+35.5}$} & \rbheat{RBHeatPurple}{30}{$\phantom{+}{0.338}$ {\fontsize{5.8}{6.5}\selectfont\textcolor{RBMuted}{$\pm 0.195$}}} & \rbheat{RBHeatPurple}{30}{${24.6}$} & \rbheat{RBHeatPurple}{31}{${42.9}$} & \rbcell{RBTeal}{15}{$+0.203$ {\fontsize{5.8}{6.5}\selectfont\textcolor{RBMuted}{$\pm 0.027$}}} & 4.8 & \rbheat{RBHeatBlue}{100}{${-7.2}\phantom{0}$} & \rbheat{RBHeatBlue}{101}{$\mathbf{-6.9}\phantom{0}$} \\
\rbdashrule
\rowcolor{black!4}
\textit{Simple logit baseline} & ${+35.4}$ & $\phantom{+}{0.343}$\phantom{ {\fontsize{5.8}{6.5}\selectfont\textcolor{RBMuted}{$\pm 0.000$}}} & ${23.8}$ & ${42.0}$ & -- & -- & ${-46.8}$ & ${-24.6}$ \\
\rbdashrule
Kimi-K3 & \rbheat{RBHeatGreen}{25}{${+32.2}$} & \rbheat{RBHeatPurple}{31}{$\phantom{+}{0.348}$ {\fontsize{5.8}{6.5}\selectfont\textcolor{RBMuted}{$\pm 0.112$}}} & \rbheat{RBHeatPurple}{31}{${24.8}$} & \rbheat{RBHeatPurple}{31}{${42.7}$} & \rbcell{RBTeal}{15}{$+0.354$ {\fontsize{5.8}{6.5}\selectfont\textcolor{RBMuted}{$\pm 0.049$}}} & 5.4 & \rbheat{RBHeatBlue}{38}{${-12.0}$} & \rbheat{RBHeatBlue}{37}{${-11.8}$} \\
Hunyuan-4-Preview & \rbheat{RBHeatGreen}{24}{${+30.5}$} & \rbheat{RBHeatPurple}{32}{$\phantom{+}{0.354}$ {\fontsize{5.8}{6.5}\selectfont\textcolor{RBMuted}{$\pm 0.202$}}} & \rbheat{RBHeatPurple}{26}{${23.2}$} & \rbheat{RBHeatPurple}{25}{${40.7}$} & $+0.018$ {\fontsize{5.8}{6.5}\selectfont\textcolor{RBMuted}{$\pm 0.008$}} & 7.8 & \rbheat{RBHeatBlue}{16}{${-32.6}$} & \rbheat{RBHeatBlue}{25}{${-17.2}$} \\
DeepSeek-V4.1-Flash & \rbheat{RBHeatGreen}{23}{${+30.0}$} & \rbheat{RBHeatPurple}{29}{$\phantom{+}{0.323}$ {\fontsize{5.8}{6.5}\selectfont\textcolor{RBMuted}{$\pm 0.130$}}} & \rbheat{RBHeatPurple}{27}{${23.8}$} & \rbheat{RBHeatPurple}{27}{${41.3}$} & $-0.068$ {\fontsize{5.8}{6.5}\selectfont\textcolor{RBMuted}{$\pm 0.017$}} & 5.8 & \rbheat{RBHeatBlue}{16}{${-32.7}$} & \rbheat{RBHeatBlue}{23}{${-18.6}$} \\
Kimi-K3.1-Pro & \rbheat{RBHeatGreen}{22}{${+27.5}$} & \rbheat{RBHeatPurple}{27}{$\phantom{+}{0.296}$ {\fontsize{5.8}{6.5}\selectfont\textcolor{RBMuted}{$\pm 0.201$}}} & \rbheat{RBHeatPurple}{31}{${24.9}$} & \rbheat{RBHeatPurple}{33}{${43.4}$} & \rbcell{RBOrange}{15}{$-0.056$ {\fontsize{5.8}{6.5}\selectfont\textcolor{RBMuted}{$\pm 0.012$}}} & 6.8 & \rbheat{RBHeatBlue}{20}{${-28.4}$} & \rbheat{RBHeatBlue}{27}{${-16.5}$} \\
DeepSeek-V4-Flash & \rbheat{RBHeatGreen}{19}{${+23.7}$} & \rbheat{RBHeatPurple}{22}{$\phantom{+}{0.242}$ {\fontsize{5.8}{6.5}\selectfont\textcolor{RBMuted}{$\pm 0.141$}}} & \rbheat{RBHeatPurple}{21}{${21.5}$} & \rbheat{RBHeatPurple}{23}{${40.0}$} & $+0.010$ {\fontsize{5.8}{6.5}\selectfont\textcolor{RBMuted}{$\pm 0.027$}} & 4.0 & \rbheat{RBHeatBlue}{14}{${-34.8}$} & \rbheat{RBHeatBlue}{16}{${-22.7}$} \\
MiMo-V2.5-Pro & \rbheat{RBHeatGreen}{19}{${+23.2}$} & \rbheat{RBHeatPurple}{18}{$\phantom{+}{0.179}$ {\fontsize{5.8}{6.5}\selectfont\textcolor{RBMuted}{$\pm 0.163$}}} & \rbheat{RBHeatPurple}{28}{${24.1}$} & \rbheat{RBHeatPurple}{28}{${41.6}$} & \rbcell{RBTeal}{15}{$+0.367$ {\fontsize{5.8}{6.5}\selectfont\textcolor{RBMuted}{$\pm 0.035$}}} & 6.8 & \rbheat{RBHeatBlue}{8}{${-44.5}$} & \rbheat{RBHeatBlue}{12}{${-25.0}$} \\
GPT-6-Luna & \rbheat{RBHeatGreen}{19}{${+22.4}$} & \rbheat{RBHeatPurple}{19}{$\phantom{+}{0.200}$ {\fontsize{5.8}{6.5}\selectfont\textcolor{RBMuted}{$\pm 0.162$}}} & \rbheat{RBHeatPurple}{14}{${18.6}$} & \rbheat{RBHeatPurple}{16}{${36.9}$} & \rbcell{RBTeal}{15}{$+0.035$ {\fontsize{5.8}{6.5}\selectfont\textcolor{RBMuted}{$\pm 0.008$}}} & 3.0 & \rbheat{RBHeatBlue}{4}{${-54.1}$} & \rbheat{RBHeatBlue}{7}{${-29.7}$} \\
GLM-5.3-Flash & \rbheat{RBHeatGreen}{18}{${+22.0}$} & \rbheat{RBHeatPurple}{26}{$\phantom{+}{0.286}$ {\fontsize{5.8}{6.5}\selectfont\textcolor{RBMuted}{$\pm 0.225$}}} & \rbheat{RBHeatPurple}{26}{${23.4}$} & \rbheat{RBHeatPurple}{30}{${42.4}$} & \rbcell{RBTeal}{15}{$+0.297$ {\fontsize{5.8}{6.5}\selectfont\textcolor{RBMuted}{$\pm 0.047$}}} & 6.4 & \rbheat{RBHeatBlue}{23}{${-24.6}$} & \rbheat{RBHeatBlue}{24}{${-17.8}$} \\
DeepSeek-V4-Pro & \rbheat{RBHeatGreen}{18}{${+21.4}$} & \rbheat{RBHeatPurple}{24}{$\phantom{+}{0.270}$ {\fontsize{5.8}{6.5}\selectfont\textcolor{RBMuted}{$\pm 0.162$}}} & \rbheat{RBHeatPurple}{19}{${21.0}$} & \rbheat{RBHeatPurple}{21}{${39.1}$} & \rbcell{RBTeal}{15}{$+0.166$ {\fontsize{5.8}{6.5}\selectfont\textcolor{RBMuted}{$\pm 0.029$}}} & 5.8 & \rbheat{RBHeatBlue}{32}{${-16.4}$} & \rbheat{RBHeatBlue}{30}{${-15.1}$} \\
GPT-5.6-Luna & \rbheat{RBHeatGreen}{18}{${+20.6}$} & \rbheat{RBHeatPurple}{20}{$\phantom{+}{0.214}$ {\fontsize{5.8}{6.5}\selectfont\textcolor{RBMuted}{$\pm 0.094$}}} & \rbheat{RBHeatPurple}{24}{${22.8}$} & \rbheat{RBHeatPurple}{26}{${41.0}$} & $-0.014$ {\fontsize{5.8}{6.5}\selectfont\textcolor{RBMuted}{$\pm 0.009$}} & 4.6 & \rbheat{RBHeatBlue}{12}{${-38.6}$} & \rbheat{RBHeatBlue}{19}{${-20.6}$} \\
MiMo-V2.6-Flash & \rbheat{RBHeatGreen}{17}{${+20.5}$} & \rbheat{RBHeatPurple}{19}{$\phantom{+}{0.202}$ {\fontsize{5.8}{6.5}\selectfont\textcolor{RBMuted}{$\pm 0.143$}}} & \rbheat{RBHeatPurple}{11}{${17.3}$} & \rbheat{RBHeatPurple}{16}{${36.8}$} & $+0.008$ {\fontsize{5.8}{6.5}\selectfont\textcolor{RBMuted}{$\pm 0.008$}} & 3.4 & \rbheat{RBHeatBlue}{6}{${-48.3}$} & \rbheat{RBHeatBlue}{8}{${-28.8}$} \\
GPT-5.6-Terra & \rbheat{RBHeatGreen}{14}{${+14.7}$} & \rbheat{RBHeatPurple}{17}{$\phantom{+}{0.170}$ {\fontsize{5.8}{6.5}\selectfont\textcolor{RBMuted}{$\pm 0.119$}}} & \rbheat{RBHeatPurple}{10}{${16.6}$} & \rbheat{RBHeatPurple}{12}{${35.0}$} & \rbcell{RBOrange}{15}{$-0.072$ {\fontsize{5.8}{6.5}\selectfont\textcolor{RBMuted}{$\pm 0.016$}}} & 2.8 & \rbheat{RBHeatBlue}{3}{${-58.0}$} & \rbheat{RBHeatBlue}{4}{${-33.9}$} \\
Gemma-4-26B-A4B & \rbheat{RBHeatGreen}{11}{${+7.8}\phantom{0}$} & \rbheat{RBHeatPurple}{11}{$\phantom{+}{0.070}$ {\fontsize{5.8}{6.5}\selectfont\textcolor{RBMuted}{$\pm 0.063$}}} & \rbheat{RBHeatPurple}{11}{${17.0}$} & \rbheat{RBHeatPurple}{11}{${34.4}$} & $-0.009$ {\fontsize{5.8}{6.5}\selectfont\textcolor{RBMuted}{$\pm 0.023$}} & 5.2 & \rbheat{RBHeatBlue}{5}{${-52.5}$} & \rbheat{RBHeatBlue}{10}{${-27.2}$} \\
\rbdashrule
\rowcolor{black!4}
\textit{Uniform} & $\phantom{+}{0.0}\phantom{0}$ & $\phantom{+}{0.000}$\phantom{ {\fontsize{5.8}{6.5}\selectfont\textcolor{RBMuted}{$\pm 0.000$}}} & ${7.1}\phantom{0}$ & ${23.4}$ & -- & -- & ${-57.8}$ & ${-26.0}$ \\
\rbdashrule
Nemotron-3.5-Lightning & \rbheat{RBHeatGreen}{8}{${-0.6}\phantom{0}$} & \rbheat{RBHeatPurple}{8}{$\phantom{+}{0.000}$ {\fontsize{5.8}{6.5}\selectfont\textcolor{RBMuted}{$\pm 0.000$}}} & \rbheat{RBHeatPurple}{2}{${7.3}\phantom{0}$} & \rbheat{RBHeatPurple}{2}{${23.4}$} & $+0.000$ {\fontsize{5.8}{6.5}\selectfont\textcolor{RBMuted}{$\pm 0.000$}} & 1.4 & \rbheat{RBHeatBlue}{3}{${-56.6}$} & \rbheat{RBHeatBlue}{12}{${-25.0}$} \\
Nemotron-3-Ultra-550B & \rbheat{RBHeatGreen}{5}{${-10.2}$} & \rbheat{RBHeatPurple}{5}{${-0.083}$ {\fontsize{5.8}{6.5}\selectfont\textcolor{RBMuted}{$\pm 0.247$}}} & \rbheat{RBHeatPurple}{7}{${14.9}$} & \rbheat{RBHeatPurple}{7}{${31.7}$} & \rbcell{RBTeal}{15}{$+0.359$ {\fontsize{5.8}{6.5}\selectfont\textcolor{RBMuted}{$\pm 0.061$}}} & 9.2 & \rbheat{RBHeatBlue}{9}{${-42.7}$} & \rbheat{RBHeatBlue}{15}{${-23.4}$} \\
Qwen3.5-35B-A3B & \rbheat{RBHeatGreen}{2}{${-25.2}$} & \rbheat{RBHeatPurple}{2}{${-0.218}$ {\fontsize{5.8}{6.5}\selectfont\textcolor{RBMuted}{$\pm 0.428$}}} & \rbheat{RBHeatPurple}{4}{${12.4}$} & \rbheat{RBHeatPurple}{4}{${28.0}$} & $-0.234$ {\fontsize{5.8}{6.5}\selectfont\textcolor{RBMuted}{$\pm 0.133$}} & 10.6 & \rbheat{RBHeatBlue}{2}{${-67.5}$} & \rbheat{RBHeatBlue}{2}{${-40.7}$} \\
Qwen3.5-9B & \rbheat{RBHeatGreen}{2}{${-33.0}$} & \rbheat{RBHeatPurple}{2}{${-0.280}$ {\fontsize{5.8}{6.5}\selectfont\textcolor{RBMuted}{$\pm 0.291$}}} & \rbheat{RBHeatPurple}{5}{${13.2}$} & \rbheat{RBHeatPurple}{4}{${28.6}$} & \rbcell{RBOrange}{15}{$-0.304$ {\fontsize{5.8}{6.5}\selectfont\textcolor{RBMuted}{$\pm 0.065$}}} & 13.0 & \rbheat{RBHeatBlue}{3}{${-60.9}$} & \rbheat{RBHeatBlue}{19}{${-20.4}$} \\
\midrule
\rowcolor{black!4}
Closing market & -- & $\phantom{+}{0.757}$\phantom{ {\fontsize{5.8}{6.5}\selectfont\textcolor{RBMuted}{$\pm 0.000$}}} & ${34.0}$ & ${51.4}$ & -- & -- & -- & ${-19.4}$ \\
\bottomrule
\end{tabular*}
\endgroup
\end{table}

\subsection{Evaluation Metrics}
\label{sec:score}
We evaluate all $N=\nRaces{}$ races per trajectory, including repaired, fallback, and uniform outputs across three decoupled dimensions:

\paragraph{Forecast quality.}
We apply the logarithmic score~\citep{gneiting2007strictly} after mixing each forecast with a 2\% uniform component. After removing scratched runners, let $p_{ri}$ be the renormalized probability over the $n_r$ actual starters, smoothed as $\widetilde p_{ri}=0.98p_{ri}+0.02/n_r$. For official winner set $\mathcal{W}_r$ of size $k_r$, the smoothed log gain relative to uniform is
\[
g_r=\ln\!\left(\frac{n_r}{k_r}\sum_{i\in \mathcal{W}_r}\widetilde p_{ri}\right).
\]
We also report \textbf{Top-1} win rate ($h_r \in \{0,1\}$ for the top-ranked surviving starter) and \textbf{Top-3} precision (share of the top $\min(3,n_r)$ surviving ranked starters finishing in the official top three).

\paragraph{Synthetic net utility.}
Following the call to evaluate cost jointly with accuracy~\citep{kapoor2024agents}, the agents' objective, which is also our ranking metric, is a net utility in which one nat of log gain per race is worth 100 score-yen, the benchmark's synthetic unit:
\begin{align*}
\mathrm{Net}&=100\sum_r\left[g_r+I_r\left(h_r-\frac{1}{n_r}\right)\right]-C, \\
C&=0.5\left(0.01N_{\mathrm{cell}}+0.5N_{\mathrm{call}}+0.2T_{\mathrm{CPU}}\right).
\end{align*}
Here, the hit term rewards top-pick accuracy with zero expectation under random ranking in races with a single official winner, gated by $I_r=1$ for valid non-uniform predictions. The research fee $C$ prices evidence cells $N_{\mathrm{cell}}$, tool calls $N_{\mathrm{call}}$, and sandbox CPU seconds $T_{\mathrm{CPU}}$ (Appendices~\ref{app:allowances}--\ref{app:tariff}), while API token costs are tracked separately. A free uniform forecast scores $0$, so negative $\mathrm{Net}$ means research costs exceed forecast gains; we report per-race $\mathrm{Net}/N$ and $C/N$.

\paragraph{Retrospective betting outcomes.}
To test whether predictive signals survive real market frictions, we settle sealed predictions retrospectively at official Japan Racing Association (JRA) dividends, which embed a $\sim$20\% parimutuel takeout; Appendices~\ref{app:kelly}--\ref{app:workflow-settlement} give the settlement rules. We consider two bet types: a \textbf{WIN} ticket pays only if the backed runner finishes first, and a \textbf{PLACE} ticket pays a smaller dividend if it finishes in the top three (top two in fields of 5--7 starters). Two sizing rules set the stake on the sealed top pick. Under \textbf{Flat} staking, each race receives JPY 100. Under \textbf{Capped Kelly}~\citep{kelly1956new}, the stake depends on the day-start bankroll $B_d$ (initial JPY 100,000), the raw probability $q_r$ of the top pick, and its last pre-race market odds $O_r$:
\[
\begin{aligned}
f_r &= \max\left\{0,\frac{q_rO_r-1}{O_r-1}\right\},\\
\text{stake} &= B_d\min\{f_r,0.01\}.
\end{aligned}
\]
Main results use WIN tickets under both rules, flat WIN and capped-Kelly WIN; Table~\ref{tab:money} reports PLACE under flat staking only. For the nine-model workflow comparison, Table~\ref{tab:workflow-betting-risk} reports mean terminal profit and loss (P\&L), turnover return on investment (ROI), maximum drawdown (MDD), and the annualized P\&L Sharpe ratio.

\section{Experiments and Analysis}
\label{sec:experiments}

All runs cover the same \nRaces{} races across \nDays{} race days between January 6 and February 3, 2024. Across \nModels{} model--scaffold configurations, \emph{code-as-policy} yields \nARuns{} \Live{}--\Frozen{} pairs. Appendix~\ref{app:models} lists model access and cohorts. For nine shared models, we compare the first three rounds of code-as-policy \Live{} and LLM-as-policy.

\subsection{Initial Feature Design and Calibration Drive Predictor Quality}
\label{sec:quality}

\paragraph{Frontier agents lead in both accuracy and utility.}
Table~\ref{tab:main} orders agents by native net utility ($\mathrm{Net}/N$) alongside log gain and Top-1/Top-3 accuracy, and Table~\ref{tab:configs} gives their fees, delivery, and scaffolds. GPT-6-Astra leads net utility and \bestModel{} leads log gain. Among the other Tier-1 configurations, cost-disciplined Gemini~4~Argon outperforms heavier spenders such as GPT-5.6-Sol. At the bottom, four configurations sit at or below uniform: Nemotron-3.5-Lightning never delivers a valid predictor, while Qwen3.5-35B-A3B and Qwen3.5-9B deliver code that runs but models the races worse than chance.

\paragraph{Feature design, not sample size, separates agents from fitted baselines.}
With 100 training races, conditional logit reaches 0.470 nats using agent-inspired features and 0.411 using 18 hand-built card, form, and connection features (Table~\ref{tab:ceiling}). The former score is slightly above the 0.462 nats achieved by the \emph{large-sample logit} using the same 18 hand-built features on 20,730 pre-window races. With broad factual features, conditional logit scores 0.394 nats at 100 races and 0.492 at 300. In a separate experiment on 87 labeled support races, the \emph{simple logit} scores 0.343 nats using race-card features and up to five recent starts. Expanding its feature set lowers its gain to 0.298 nats, as Table~\ref{tab:refs} shows. In Table~\ref{tab:main}, all nine Tier-1 configurations outperform the simple logit, and six outperform the large-sample logit. The top predictors combine feature engineering with probability calibration. GPT-6-Astra standardizes 42 features before computing a linear score, while GPT-6.1-Sol centers 86 features within each race and blends a linear score with depth-two trees. Both mix 2--3\% uniform probability mass into their forecasts. Appendix~\ref{app:case-astra} gives the full case study.

\paragraph{Output audits find no evidence of result recall.}
We audited agent messages, tool arguments, and saved predictors across
431 code-as-policy \Live{} trajectories. With identifiers pseudonymized
and dates represented as relative day indices, we found no instance of
an agent mapping a day index to a calendar date or recalling a race result. Appendix~\ref{app:priors} describes the audit procedure and the remaining possibility of race re-identification.

\subsection{Revision Repairs Weak Predictors but Rarely Improves Strong Ones}
\label{sec:revision}

Given that the top initial predictors already excel, when does online revision actually help?

\begin{figure}[!t]
\centering
\includegraphics[width=\textwidth]{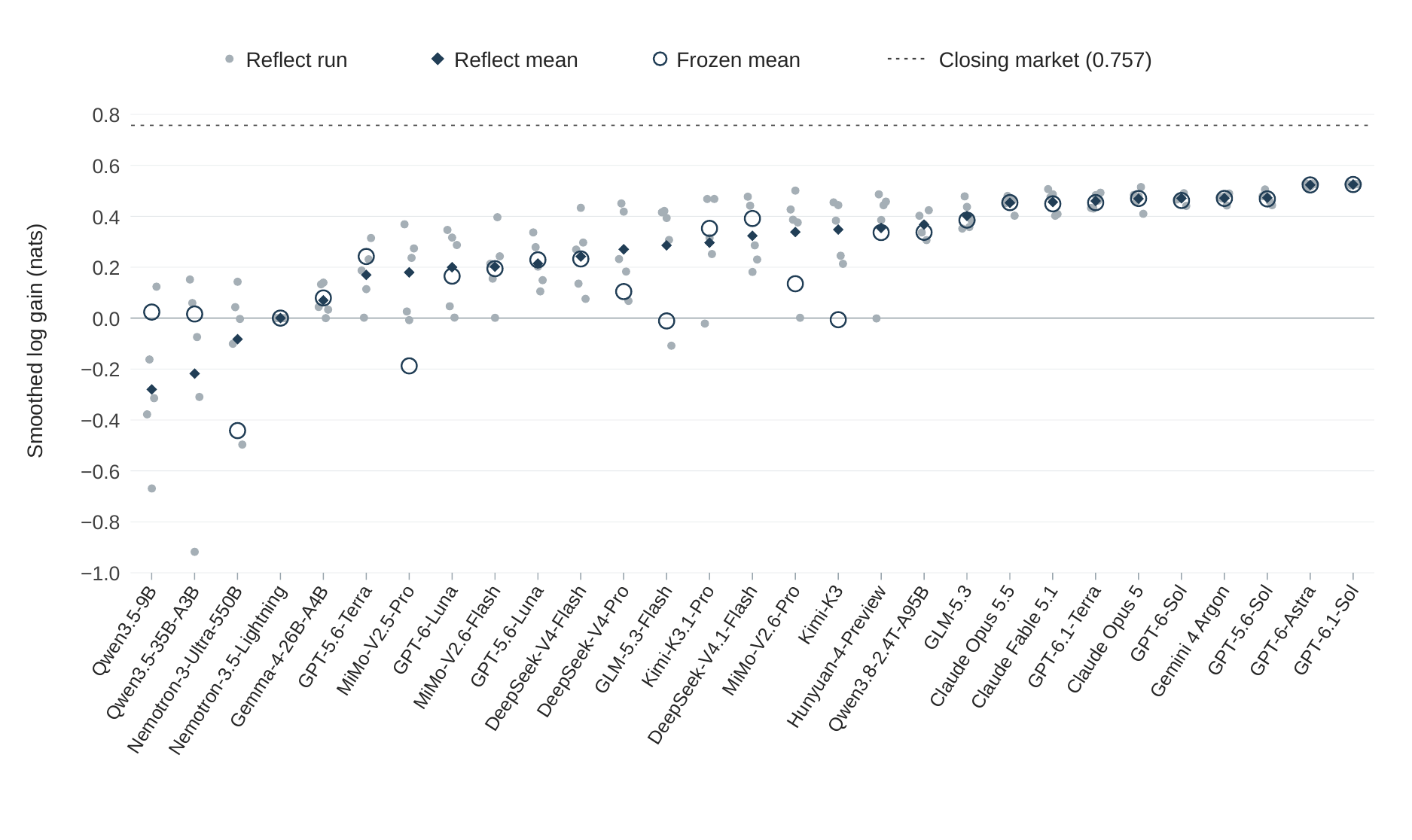}
\caption{Code-as-policy forecast quality for all \nModels{} configurations in Table~\ref{tab:main}, ordered by increasing mean \Live{} smoothed log gain. Gray dots are the five individual \Live{} trajectories per configuration (\nARuns{} in total); diamonds are their means, and open circles are the means of the five paired \Frozen{} trajectories. Horizontal jitter separates overlapping dots. The solid zero line denotes the uniform reference, and the dotted line marks the closing market's log gain (0.757 nats), computed from odds withheld from agents. Every trajectory is evaluated on the same \nRaces{} races across \nDays{} days, so the spread describes between-run variation on a shared outcome window. Statistical inference follows the paired day-clustered tests in Table~\ref{tab:main}.}
\label{fig:code-policy-forecast-quality}
\end{figure}

\paragraph{Strong initial predictors are never revised.}
The best agents barely touch their first predictor. Table~\ref{tab:main} groups configurations by net-utility rank: Tier~1 comprises ranks 1--9, Tier~2 ranks 10--12, Tier~3 ranks 13--25, and Tier~4 ranks 26--29. None of the nine Tier-1 configurations shows a significant \Live{}$-$\Frozen{} change in either direction; they use at most five predictor versions per run, and the top two, GPT-6-Astra and \bestModel{}, never change theirs at all. Figure~\ref{fig:code-policy-forecast-quality} shows all five \Live{} runs and paired \Frozen{} means. Leaving the predictor alone is the right call: a day releases only 24--36 results, too few to separate a real weakness from a bad day, and these predictors are calibrated when written: \bestModel{} scales its score before the softmax and mixes in 3\% uniform mass, as Appendix~\ref{app:case-sol} shows. Their first predictors set a high floor that no revised predictor below them reaches.

\paragraph{Below Tier 1, revision repairs flaws or chases noise.}
Among the remaining configurations, \nHolmPos{} improve and three decline significantly after Holm correction~\citep{holm1979simple}. Gains come from repairing a flawed $P_0$, chiefly overconfidence. The initial predictor of Kimi-K3 applies a temperature-1 softmax to 16 fixed weights and loses 0.243 nats on day 1; raising the temperature to 2--3 changes no top pick but cuts mean top-pick probability from 53.9\% to 29.3\%, lifting this run from 0.093 to 0.383 nats, as Appendix~\ref{app:case-kimi} shows. MiMo-V2.5-Pro and Nemotron-3-Ultra-550B repair it the same way and gain $0.367$ and $0.359$ nats over their \Frozen{} controls. Declines come from acting on a biased or noisy signal. GPT-5.6-Terra drops recent form on day 4 to save fees, after its own warmup test, run on the oldest archive races where almost no runner has form, made form look worthless; its log gain falls from 0.333 to 0.187 nats, as Appendices~\ref{app:case-terra} and~\ref{app:revision-results} show. Kimi-K3.1-Pro and Qwen3.5-9B instead rewrite their features day after day, the latter through 13 predictors per run, and lose $0.056$ and $0.304$ nats by fitting the noise of each day, like DeepSeek-V4.1-Flash, whose feature count moves 12, 7, 10, 14, 25, as Appendix~\ref{app:case-deepseek} shows.

\subsection{Forecast Quality and Research Cost across Workflows}
\label{sec:code-vs-llm-policy}

\begin{table}[!t]
\caption{\textbf{Code-as-policy vs. LLM-as-policy.} The first two rows average three runs per model across nine models. The last row counts model-level wins. Net uses the common workflow rescoring defined in Appendix~\ref{app:acquisition-scoreyen}. Per-model results are in Tables~\ref{tab:workflow-forecast-quality}, \ref{tab:workflow-betting-risk} and~\ref{tab:research-net-scores}.}
\label{tab:workflow-summary}
\centering\footnotesize
\setlength{\tabcolsep}{3pt}
\begin{tabular}{@{}lccccc@{}}
\toprule
Workflow & \shortstack[c]{Log\\gain} & \shortstack[c]{Top-1\\(\%)} & \shortstack[c]{Fee\\/race} & \shortstack[c]{Net\\/race} & \shortstack[c]{Kelly P\&L\\(JPY)} \\
\midrule
\CodePolicyWorkflow & \textbf{0.392} & \textbf{25.5} & \textbf{15.9} & \textbf{41.2} & $\mathbf{-17{,}737}$ \\
\LLMPolicyWorkflow & 0.316 & 21.9 & 25.6 & 20.5 & $-40{,}192$ \\
\midrule
\shortstack[l]{\CodePolicyWorkflow\\better on} & 7/9 & 8/9 & 9/9 & 8/9 & 7/9 \\
\bottomrule
\end{tabular}
\end{table}

\begin{figure}[!t]
\centering
\includegraphics[width=\textwidth]
{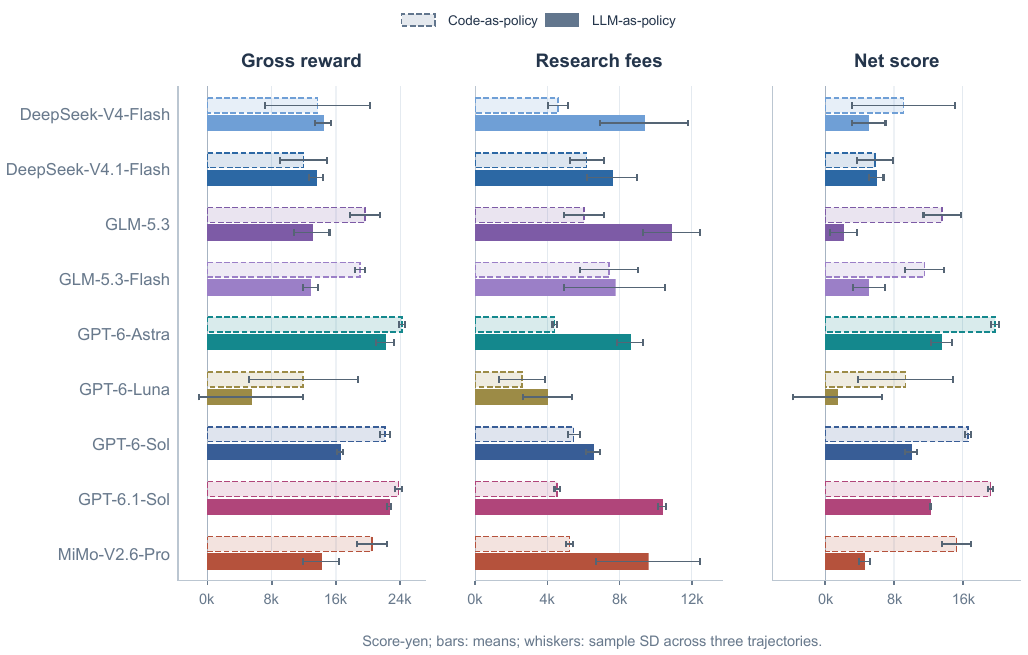}
\caption{Mean trajectory gross reward, weighted research fees, and net utility by workflow ($n=3$; whiskers show between-trajectory sample SDs).}
\label{fig:workflow-research-scores}
\end{figure}

Having examined agents that compile reasoning into code, we next compare this paradigm with invoking the LLM on every race across the nine-model workflow cohort.

\paragraph{Code-as-policy delivers higher forecast quality.}
Across the nine shared models, code-as-policy achieves higher mean log gain and higher mean top-pick accuracy than LLM-as-policy. Table~\ref{tab:workflow-summary} shows that code-as-policy leads on log gain for seven models, averaging 0.392 nats versus 0.316, and on top-pick accuracy for eight, averaging 25.5\% versus 21.9\%. Table~\ref{tab:workflow-forecast-quality} shows particularly large log-gain differences for GPT-6-Sol, at 0.474 versus 0.367 nats, and MiMo-V2.6-Pro, at 0.438 versus 0.281. The two workflows also differ in their evidence interfaces, so these results reflect both how evidence is acquired and how forecasts are produced.

\paragraph{LLM-as-policy improves forecasts when model-written predictors are unreliable.}
Under LLM-as-policy, the saved program collects evidence and the LLM estimates win probabilities for each race. Appendix~\ref{app:workflow-forecast-quality} shows that 88.7--99.8\% of its forecasts are direct. For DeepSeek-V4.1-Flash, LLM-as-policy leads on all three forecast metrics, with log gain rising from 0.233 to 0.277 nats. For DeepSeek-V4-Flash, log gain rises from 0.269 to 0.297 nats.

\paragraph{Recent-history retrieval accounts for most LLM-as-policy acquisition fees.}
Under the same fee schedule, code-as-policy incurs lower research fees for all nine models and achieves higher net utility for eight. Table~\ref{tab:workflow-summary} shows that fees average 15.9 score-yen per race for code-as-policy and 25.6 for LLM-as-policy. The breakdown in Appendix~\ref{app:acquisition-by-model} attributes 92.9\% of LLM-as-policy acquisition fees to recent-history retrieval. Figure~\ref{fig:workflow-research-scores} shows the per-model breakdown in synthetic score-yen.

\paragraph{The two highest-scoring models keep their Day~2 evidence plans.}
Across all six LLM-as-policy runs of GPT-6.1-Sol and GPT-6-Astra, the plan adopted on Day~2 remains unchanged through Day~10. Table~\ref{tab:plan-cadence} shows that each of the other 21 runs revises its plan at least three times. Weaker early forecasts may themselves prompt further revision.

\subsection{Forecast Gains Do Not Survive Takeout}
\label{sec:workflow-betting}

\paragraph{No configuration profits after takeout.}
\label{sec:money}
Despite positive log gains for 25 of \nModels{} configurations, every configuration has a negative mean return across the five-run cohort under both 1\%-capped Kelly ($-3.2\%$ to $-67.5\%$) and flat WIN ($-6.9\%$ to $-40.7\%$), and only 13 of \nARuns{} individual Kelly runs end in profit. Table~\ref{tab:main} and Figures~\ref{fig:teaser} and~\ref{fig:equity} show these returns, and Appendix~\ref{app:return-results} breaks them down. Eighteen of the \nModels{} configurations lose less under flat WIN than backing the closing favorite ($-19.4\%$), but beating a uniform prior is far from enough to overcome the $\sim20\%$ parimutuel takeout.\looseness=-1

\begin{figure}[!t]
\centering
\includegraphics[width=\textwidth]{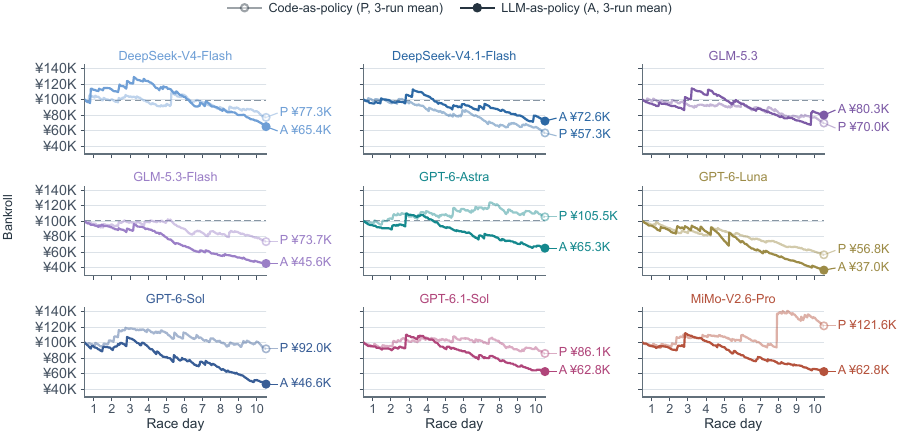}
\caption{\textbf{Mean bankroll under 1\%-capped Kelly staking.} For each of the nine models, curves show the mean across rounds 1--3. Dashed \texttt{P}: code-as-policy. Solid \texttt{A}: LLM-as-policy. Initial bankroll: JPY~100,000 per run.}
\label{fig:workflow-betting-profit}
\end{figure}

\paragraph{Code-as-policy loses less than LLM-as-policy.}
Under the shared settlement rules of Appendix~\ref{app:workflow-settlement}, withdrawals are refunded and Kelly stakes use the last pre-race odds snapshot (TYB). Table~\ref{tab:workflow-betting-risk} shows that code-as-policy achieves higher mean terminal P\&L on eight of nine models under flat WIN and seven under capped Kelly, with smaller mean-path drawdowns on nine and eight, respectively. Figure~\ref{fig:workflow-betting-profit} shows the capped-Kelly bankroll paths, and Figure~\ref{fig:workflow-flat-profit} shows cumulative profit under flat WIN for the same models. All nine LLM-as-policy means are negative under both rules.

\paragraph{Odds-blind selection induces a winner's curse.}
Why do models with positive log gain still lose under Kelly staking? Agents estimate $q_r$ without seeing the odds $O_r$, so the trigger $q_rO_r>1$ favors races where the estimate of the agent is high relative to the market, which includes those where it is overoptimistic; the 1\% cap binds on 96\% of placed bets in the main cohort, so capped Kelly effectively acts as a fixed-fraction stake on those races (Appendix~\ref{app:return-results}). Although LLM-as-policy underpredicts win rates overall on seven of nine models, both workflows overpredict on their Kelly-selected bets for all nine, a pattern consistent with the optimizer's curse~\citep{smith2006optimizer} that Appendix~\ref{app:bet-selection} examines. For DeepSeek-V4-Flash LLM-as-policy, predicted versus observed win rates move from $17.94\%$ vs.\ $22.05\%$ on all races to $17.47\%$ vs.\ $7.16\%$ on selected bets.

\paragraph{Few runs finish in profit.}
Table~\ref{tab:workflow-betting-risk} shows that only 8 of 54 workflow runs finish in profit under capped Kelly. For GPT-6-Astra and MiMo-V2.6-Pro, mean code-as-policy profit is positive over the first three runs but turns negative when all five runs are included, as Table~\ref{tab:workflow-betting-risk} and Figure~\ref{fig:teaser} show.

\FloatBarrier
\section{Conclusion}
Sequential probabilistic forecasting confronts agents and evaluators with the same obstacle: realized outcomes are noisy, making it hard to attribute performance to true forecasting skill rather than chance. In \textsc{JockeyBench}, the strongest agents draw their edge from the feature design and calibration of their first predictor and leave it unchanged; below them, revision helps when it repairs a clear flaw such as overconfidence, but hurts when it chases daily noise. For the evaluator, settled bets compound this noise with the 20\% parimutuel takeout: 25 of \nModels{} configurations beat a uniform forecast in log gain, yet only 13 of \nARuns{} capped-Kelly runs end in profit. We therefore decouple forecast quality from realized profit, scoring probabilities with a proper scoring rule and isolating the effect of revision against a frozen copy of the first predictor via replay. This lens also clarifies the trade-off between frameworks: code-as-policy is more accurate and cheaper for most models while yielding a reusable predictor that can be frozen and replayed, whereas LLM-as-policy helps when a model's own predictors are unreliable. More broadly, benchmarks where agents learn from realized events should separate forecast quality, counterfactual revision gain, and realized payoff, so that noisy outcomes mislead neither the agent's updates nor the evaluator's conclusions.%
\label{end:body}

\section*{Limitations}
\label{sec:limitations}
Our evaluation covers structured records from Japanese horse racing over a \nRaces{}-race window across ten race days. Although \textsc{JockeyBench} seals outcomes and pseudonymizes records, historical races remain re-identifiable in principle, so our output audit and schema-blinding checks (Appendix~\ref{app:priors}) provide empirical evidence rather than a formal guarantee against memorization. The window is also short for settled returns. A few winning tickets can dominate a run's payout, so returns at this scale cannot separate models, and we therefore rank models by forecast quality.

In addition, the two workflows differ in both prediction mechanism and evidence interface, and each main-cohort model runs in its native scaffold. The workflow comparison therefore measures complete systems rather than either factor alone, and deployment cost depends on provider pricing and routing~\citep{irassar2026you}. Agents also never see market odds. In a matched odds-visible arm, they converge on a calibrated market price (Appendix~\ref{app:odds}), so whether agents can add information beyond that price remains open.

In future work, we will pursue two directions. First, we will extend \textsc{JockeyBench} into a prospective living benchmark on future race meetings. Future races cannot be memorized, and a multi-season horizon would provide enough races to compare settled returns. Second, we will run shared-harness ablations that vary one factor at a time, namely the prediction mechanism, the evidence interface, the utility score, and odds visibility, to identify which design choices drive forecast quality and when agents can improve on the market price.

\section*{Ethics Statement}
Horseracing involves broader animal-welfare and social considerations~\citep{mcmanus2012global}. This study conducts retrospective simulation on historical records; no live bets were placed, score-yen is a synthetic benchmark unit, and betting metrics are computed purely as historical counterfactuals. Source records were accessed under a commercial JRDB membership. Raw archive files and private HMAC identifier mappings remain restricted to the evaluation host, and public release includes our code, synthetic verification suite, and aggregate audit manifests subject to archive redistribution terms.\looseness=-1

\bibliography{references}
\appendix
\clearpage
\setcounter{dbltopnumber}{3}
\setcounter{totalnumber}{6}
\renewcommand{\dbltopfraction}{0.95}
\renewcommand{\textfraction}{0.05}
\section*{Guide to the appendix}
\noindent\begin{tabular}{@{}p{0.84\columnwidth}r@{}}
\textbf{Contents} & \textbf{Page} \\
\midrule
\hyperref[app:implementation]{A. Benchmark and experimental details} & \pageref{app:implementation} \\
\hyperref[app:extended-results]{B. Extended results and diagnostics} & \pageref{app:extended-results} \\
\hyperref[app:prompt]{C. Prompt templates and submission interfaces} & \pageref{app:prompt} \\
\hyperref[app:case-studies]{D. Case studies of executed strategies} & \pageref{app:case-studies} \\
\hyperref[app:repro]{E. Reproducibility and released materials} & \pageref{app:repro} \\
\end{tabular}

\section{Benchmark and Experimental Details}
\label{app:implementation}
\subsection{Data, preprocessing, and information timing}
\label{app:data}

\begin{table}[!htbp]
\caption{Fields available to the agent and its predictor. Each card also carries the entrant number and pseudonymous horse, jockey and trainer IDs; categorical codes come without a codebook, and unknown values are \texttt{null} with a reason.}
\label{tab:fields}
\centering
\small
\begin{tabular}{@{}lcp{0.76\textwidth}@{}}
\toprule
Group & Fields & Content\\
\midrule
Race & 23 & distance; surface; direction; course layout and variant; venue; meeting day; age class; eligibility condition; weight rule; grade; declared field size; turf grass type and height; rolling and antifreeze treatment; rainfall; forecast weather; forecast turf and dirt going; inner, middle and outer turf condition\\
\midrule
Runner & 25 & carried weight; draw; sex; age in days; blinkers; body weight and its change (when on the card); layoff duration; transport; earned and class prize money; starts and days since stabling; class-drop, rest-reason and hoof codes; late-start rate; sire and damsire turf and dirt top-two rates and average distance; sire and damsire IDs\\
\midrule
Workout & 13 & workout distance; early, middle and late split times; effort and urging; rider role; training partner's result, class, age and effort; course type; distance type\\
\midrule
Past start & 30 & finishing position; abnormal result; distance; surface; going; venue; condition; grade; age class; field size; draw; carried weight; body weight and its change; first and last three-furlong times; margin to the winner; fourth-corner position and path; path; overall, early, middle and late trouble; race pace and own pace; running style; slow start; jockey and trainer IDs\\
\midrule
Connections & 6 & jockey and trainer starts, wins and top-three finishes over the preceding 365 days, counted from released results\\
\midrule
Withheld & -- & odds and popularity (current and past races); dividends; expert indices, ratings and marks; predicted positions; aptitude, class and track-bias assessments; free text; real names; calendar dates in race records\\
\bottomrule
\end{tabular}

\end{table}

\paragraph{Archive provenance.}
The archive is distributed as fixed-width Shift-JIS text: 40,974 daily files and annual packs (6.45\,GB) in 41 file series. Each file is stored verbatim with its SHA-256 hash and capture time, and 1,100 field definitions were transcribed from the published format specifications~\citep{jrdbDataSpecifications}, including older layout variants. Records are sliced by byte position before CP932 decoding. The 7.5 million source records in the cleaned database each trace back to a hashed file. Jockey and trainer links follow each race's declarations rather than a later master snapshot, and pseudonyms are keyed HMACs whose mapping stays on the evaluation host.

\subsection{Agent workflows and execution}
\label{app:workflows}
Both workflows follow the historical release clock in Figure~\ref{fig:data-cleaning}; reflection runs after the 20:30 outcome release on day $d$ and before the next 21:00 cutoff for day $d+1$.

\subsubsection{Code-as-policy}
The LLM constructs an initial predictor and may revise it after each released day. Race forecasts are produced by the saved predictor. The phase prompts and submission example are in Appendix~\ref{app:prompt}.

\paragraph{Interface, validation, and repair.}
\label{app:validation}
Predictors implement \texttt{predict(sdk, context)}, query data through \texttt{sdk.call}, and return complete rankings and win probabilities. Validation checks Python syntax, signature, entrant coverage, finite probabilities in $[0,1]$ summing to $1\pm0.02$, execution limits, and prohibited file writes. Candidates run twice on up to three initial-support races. Nondeterministic or exactly uniform outputs prompt repair; these flags alone neither reject predictors nor select them by accuracy. Invalid candidates retain the previous valid version; execution failures fall back to an earlier valid predictor, then deterministic uniform output. All delivery categories enter the scores.

In Table~\ref{tab:configs}, \emph{races by valid predictor} counts \texttt{predictor} deliveries as a percentage,
excluding exactly uniform outputs. Other labels are \texttt{silent\_uniform}/\texttt{explicit\_uniform}
(uniform outputs), \texttt{no\_strategy} (missing predictor), and \texttt{execution\_fallback}
(runtime fallback). In Table~\ref{tab:main}, \emph{Predictors used} averages accepted predictors per \Live{} trajectory,
including the initial predictor.

\paragraph{Allowances, persistence, and sandbox.}
\label{app:allowances}
Initial sessions allow 35 model responses plus five reserved for repair and 400,000 cells. Revision sessions allow 15 plus five repair responses and 100,000 cells. Each race execution permits 40 data calls, 5,000 cells, and 120 CPU-seconds. Repair uses the reserved responses and any unused main-session responses within the original phase ceiling, with at most one repair session per phase. Data allowances are shared within construction or revision phases, including repair; each prediction and validation execution has its own allowance.

Source, fitted parameters, concise notes, the ledger, and validation feedback persist between otherwise fresh session contexts. Historical samples and training arrays may not be persisted on disk; archive access remains mediated by the SDK. The sandbox has no network or subprocess access and fixes the interpreter, allowed libraries, thread count, and hash seed.

\subsubsection{LLM-as-policy}
\label{app:active-research}
The LLM builds an initial evidence plan, may update it after each released day, and forecasts each race in a fresh session from the evidence the plan collects. The prompts for these sessions are in Appendix~\ref{app:prompt-llm-policy}.

\paragraph{Evidence plan and interface.}
An executable evidence plan defines \texttt{collect\_evidence(sdk, context)} and runs in a restricted Python interpreter. Context contains the race identifier, day index, feature cut-off, field size, and entrant numbers. Through \texttt{sdk.state()}, the plan observes query counts, costs, domain coverage, returned-record counts, missingness, freshness, and the availability of further bounded history. The trusted broker retains raw observations and constructs the evidence package. All evidence requests use the race day's shared cutoff of 21:00 JST on the preceding day. Race outcomes become available at their scheduled release. The plan permits bounded control flow and approved SDK operations, but no external files, network access, arbitrary code execution, model calls, or access to unreleased outcomes. It returns the broker-managed evidence package rather than rankings or probabilities.

\paragraph{Forecast delivery.}
The evidence plan runs without an LLM call. A fresh forecasting session then receives the basic card and evidence package, with only the native submission tool available. A valid sealed forecast ranks all entrants and assigns each a finite,
nonnegative win probability, with the probabilities summing to one.
After a recoverable delivery error, the framework allows at most one
repair request. If the repair fails and integrity checks pass, the
framework seals a uniform forecast. Integrity failures halt the run.
All sealed forecasts are scored. Delivery records distinguish direct
and repaired non-uniform forecasts, model-submitted uniform forecasts,
and framework-generated uniform fallbacks.

\paragraph{Warmup and persistence.}
Historical warmup uses legally released support preceding the window. Its analysis can affect query order, thresholds, branches, and stopping conditions in the saved plan. The persisted artifact consists of validated source, its identity and provenance, and construction metadata. It contains no fitted predictive parameters. Raw historical records, the warmup conversation, and the analysis workspace are not transferred into the formal trajectory.

\paragraph{Updates and validation.}
After outcome release, the updater receives the current plan, current-day statistics, cumulative acquisition and cost summaries, delivery and forecast-quality measures, the change from the previous day, up to two earlier released Experience Frames, and up to three recent plan changes. It inspects the evidence-retrieval contract and either validates and saves an exact candidate or retains and validates the current source. Invalid candidates leave the previous valid plan unchanged. Validation repeatedly executes synthetic cards to check the interface, determinism, and evidence-package construction, without real-race outcomes or accuracy-based selection. Queries, stopping decisions, packages, and hashes are recorded alongside adopted evidence plans.

\subsection{Models, configurations, and evaluation cohorts}
\label{app:models}
The main cohort spans 11 model families: GPT~\citep{openaiModelDocumentation}, Claude~\citep{anthropicModelDocumentation}, Gemini~\citep{googleGeminiArgonModelCard}, GLM~\citep{zaiModelReleases}, MiMo~\citep{xiaomiModelReleases}, Qwen~\citep{qwenModelReleases}, Kimi~\citep{moonshotKimiModelCard}, DeepSeek~\citep{deepseekModelReleases}, Gemma~\citep{googleGemmaModelCard}, Nemotron~\citep{nvidiaModelReleases}, and Hunyuan~\citep{tencentHyModelRepository}. OpenAI configurations use the Codex app-server (with dynamic tools and code-mode batching), Claude configurations use Claude Code (via an MCP bridge with built-in tools disabled), Gemini 4 Argon uses Google's native GenerateContent API with the same tool contract, and all other configurations use DeepInfra's OpenAI-compatible API loop. Table~\ref{tab:configs} lists model identifiers, routes, requested reasoning effort, mean model responses per trajectory, research fees, and predictor delivery for all \nModels{} configurations.
\begin{table}[!tbp]
\caption{Configurations in the main code-as-policy cohort, in the order of Table~\ref{tab:main}. All configurations request high reasoning effort except Hunyuan-4-Preview, which sends no effort field. \emph{LLM calls per run} is the number of model responses in one \Live{} trajectory, across warmup and daily revision; data-tool calls may be batched differently by route. \emph{Research fee per race} is the score-yen charged for data, calls, and compute, divided by 324 races. \emph{Races by valid predictor} is the share of races forecast by a validated predictor whose output is not uniform; the remaining races received uniform output, no predictor, or a runtime fallback (Appendix~\ref{app:validation}). Every configuration has five complete 324-race \Live{}/\Frozen{} pairs (\nARuns{} in total); all values are five-run means. Model IDs retain the recorded provider aliases.}
\label{tab:configs}
\centering
\renewcommand{\arraystretch}{1.1}\footnotesize
\setlength{\tabcolsep}{3pt}
\begin{tabular*}{\textwidth}{@{\extracolsep{\fill}}lccccc@{}}
\toprule
Display name & Model identifier & Route & \shortstack[c]{LLM\\calls\\per run} & \shortstack[c]{Research\\fee\\per race} & \shortstack[c]{Races by\\valid\\predictor (\%)} \\
\midrule
GPT-6-Astra & gpt-6-astra & Codex & 25.8 & 13.6 & 100.0 \\
GPT-6.1-Sol & gpt-6.1-sol & Codex & 38.0 & 14.1 & 100.0 \\
Gemini 4 Argon & gemini-4-argon & API & 42.0 & 12.8 & 100.0 \\
Claude Opus 5.5 & claude-opus-5-5 & Claude Code & 30.4 & 12.2 & 100.0 \\
GPT-6.1-Terra & gpt-6.1-terra & Codex & 28.2 & 12.8 & 100.0 \\
Claude Opus 5 & claude-opus-5 & Claude Code & 48.8 & 15.5 & 100.0 \\
Claude Fable 5.1 & claude-fable-5-1 & Claude Code & 51.6 & 14.1 & 100.0 \\
GPT-5.6-Sol & gpt-5.6-sol & Codex & 78.6 & 17.0 & 100.0 \\
GPT-6-Sol & gpt-6-sol & Codex & 95.6 & 16.7 & 100.0 \\
GLM-5.3 & zai-org/GLM-5.3 & API & 60.6 & 18.7 & 100.0 \\
Qwen3.8-2.4T-A95B & Qwen/Qwen3.8-2.4T-A95B & API & 65.0 & 16.1 & 100.0 \\
MiMo-V2.6-Pro & XiaomiMiMo/MiMo-V2.6-Pro & API & 58.0 & 15.5 & 100.0 \\
Kimi-K3 & moonshotai/Kimi-K3 & API & 56.6 & 19.9 & 100.0 \\
Hunyuan-4-Preview & tencent/Hy4-preview & API & 119.4 & 20.0 & 86.7 \\
DeepSeek-V4.1-Flash & deepseek-ai/DeepSeek-V4.1-Flash & API & 115.2 & 18.6 & 100.0 \\
Kimi-K3.1-Pro & moonshotai/Kimi-K3.1-Pro & API & 83.6 & 19.5 & 100.0 \\
DeepSeek-V4-Flash & deepseek-ai/DeepSeek-V4-Flash-0731 & API & 63.8 & 14.5 & 100.0 \\
MiMo-V2.5-Pro & XiaomiMiMo/MiMo-V2.5-Pro & API & 73.0 & 11.4 & 99.8 \\
GPT-6-Luna & gpt-6-luna & Codex & 26.4 & 8.6 & 95.1 \\
GLM-5.3-Flash & zai-org/GLM-5.3-Flash & API & 85.4 & 22.5 & 100.0 \\
DeepSeek-V4-Pro & deepseek-ai/DeepSeek-V4-Pro-0813 & API & 106.8 & 19.1 & 99.9 \\
GPT-5.6-Luna & gpt-5.6-luna & Codex & 55.0 & 16.1 & 100.0 \\
MiMo-V2.6-Flash & XiaomiMiMo/MiMo-V2.6-Flash & API & 56.8 & 9.6 & 98.3 \\
GPT-5.6-Terra & gpt-5.6-terra & Codex & 51.8 & 11.4 & 98.1 \\
Gemma-4-26B-A4B & google/gemma-4-26B-A4B-it & API & 29.6 & 8.5 & 70.0 \\
Nemotron-3.5-Lightning & nvidia/NVIDIA-Nemotron-3.5-Lightning & API & 54.8 & 0.6 & 0.0 \\
Nemotron-3-Ultra-550B & nvidia/NVIDIA-Nemotron-3-Ultra-550B-A55B & API & 78.0 & 9.3 & 98.4 \\
Qwen3.5-35B-A3B & Qwen/Qwen3.5-35B-A3B & API & 62.2 & 8.5 & 89.1 \\
Qwen3.5-9B & Qwen/Qwen3.5-9B & API & 100.2 & 10.4 & 69.6 \\
\bottomrule\end{tabular*}
\end{table}

\paragraph{Workflow comparison and supplementary cohorts.}
The balanced dual-workflow comparison covers nine shared models---DeepSeek-V4-Flash, DeepSeek-V4.1-Flash, GLM-5.3, GLM-5.3-Flash, GPT-6-Astra, GPT-6-Luna, GPT-6-Sol, GPT-6.1-Sol, and MiMo-V2.6-Pro---using code-as-policy rounds 1--3 from the main cohort and LLM-as-policy rounds 1--3 (54 trajectories total, 3 per model and workflow). Appendix~\ref{app:matched-frozen-acquisition} reports five supplementary
pairs with frozen and revisable evidence plans, covering four DeepSeek-V4-Flash
runs (R1, R2, R4, and R6) and one GPT-6-Astra run (R1). The LLM-as-policy acquisition analysis (Appendix~\ref{app:acquisition-by-model}) uses the 27 LLM-as-policy trajectories of this cohort, and the return-concentration analysis (Appendix~\ref{app:deepseek-return-concentration}) uses all six completed DeepSeek-V4-Flash LLM-as-policy repeats (R1--R4, R6, R7). Seven of the 27 LLM-as-policy trajectories resumed from saved checkpoints without altering sealed forecasts, and three required aggregate reconstruction after native terminal-report failures; sensitivity checks excluding resumed runs, reconstructed reports, and the probability-normalized GLM-5.3-Flash R2 run leave all workflow-comparison directions unchanged.








\subsection{Forecasting references}
\label{app:refs}
\begin{table}[!tbp]
\caption{Forecasting references on the main evaluation window (324 races, 6 January--3 February 2024). All rows are scored on the same 324 races; the training samples named in the group rows precede the window. Log gain is in nats relative to uniform; Top-1 is the win rate of top-ranked non-scratched runners (0--1). $\uparrow$: higher is better. R0 reports empirical accuracy under a seeded pseudorandom ranking. Agent-inspired features were selected post hoc. Methods: Appendix~\ref{app:refs}.}
\label{tab:refs}
\centering
\renewcommand{\arraystretch}{1.04}
\footnotesize
\setlength{\tabcolsep}{2pt}
\begin{tabular*}{\columnwidth}{@{\extracolsep{\fill}}lccc@{}}
\toprule
Reference & \shortstack[c]{Log\\gain $\uparrow$} & \shortstack[c]{Top-1\\acc. $\uparrow$} & Predictor \\
\midrule
\multicolumn{4}{l}{\textbf{Agent interface} (trained on 87 support races)} \\
R0: uniform & 0.000 & 0.071 & Equal probabilities \\
R1: race card & 0.155 & 0.157 & Cond.\ logit \\
R2: card + recent history & 0.343 & 0.238 & Cond.\ logit \\
R3: expanded factual & 0.298 & 0.210 & Cond.\ logit \\
\midrule
\multicolumn{4}{l}{\textbf{Offline} (trained on 20,730 pre-window races)} \\
18 hand-built & 0.462 & 0.253 & Cond.\ logit \\
Agent-inspired & 0.523 & 0.278 & Logit + trees \\
Broad factual & 0.608 & 0.309 & Logit + trees \\
\midrule
\multicolumn{4}{l}{\textbf{External} (inputs unavailable to agents)} \\
Expert forecast & 0.679 & 0.318 & Calib.\ inverse odds \\
Pre-race market & 0.700 & 0.318 & Norm.\ inverse odds \\
Closing market & 0.757 & 0.340 & Norm.\ inverse odds \\
\bottomrule
\end{tabular*}
\end{table}

All reference models output runner win probabilities and rankings on the main evaluation window. Identifiers R0--R3 denote support-fitted reference predictors using the agent interface, while F18, FAGENT, and FV5 denote offline feature sets. The references are scored on the same 324 races as both workflows. Uniform and the simple logit run under the agents' tariff; the large-sample logit is trained offline outside it, so Table~\ref{tab:main} reports no net utility for it. Within each workflow, the paired control is instead the agent's own frozen initial artifact: the predictor $P_0$ in code-as-policy (Table~\ref{tab:main}) and the evidence plan $P_0$ in LLM-as-policy (Appendix~\ref{app:matched-frozen-acquisition}).

\paragraph{References using the agent interface.}
R0 outputs uniform probabilities (with Top-1 evaluated under a seeded pseudorandom ranking). R1 (\emph{race-card features}) uses nine card measurements: two log-transformed prize totals, carried weight, layoff weeks, age, draw, starts since stabling, and two workout times, standardized within each race with missing-value indicators. R2 (\emph{card and recent history}) adds average and best finish, win and top-three rates, and usable start counts over up to five prior starts. R3 (\emph{expanded factual features}) adds horse-condition, pedigree, workout, race-condition, and connection fields averaged over prior starts, with string categories hashed to numeric values. R1--R3 fit the same $\ell_2$-regularized conditional logit on 87 labeled support races under the agent tariff, keeping weights fixed during evaluation while refreshing pre-race inputs.

\paragraph{Offline feature construction.}
F18 (18 hand-built features) combines card and workout measurements with recent finish percentiles, career record, recency, jockey/trainer win rates, and a no-history indicator. FAGENT (\emph{agent-inspired features}, curated post hoc after inspecting top agent predictors) combines recent, best, and recency-weighted finishes, transformed finish margins, race-condition changes, and smoothed connection rates. FV5 (\emph{broad factual features}) aggregates summaries over the last 1, 3, and 5 starts, workout metrics, and connection/pedigree statistics, encoding the 16 most frequent categorical levels alongside other and missing indicators. None of these representations uses market odds or expert ratings.

\paragraph{Offline fitting and training data.}
Offline references train on 20,730 historical races (6 January 2018--28 December 2023), split chronologically by day into 80\% fitting, 10\% selection, and 10\% calibration blocks (16,578 / 2,076 / 2,076 races). Conditional logit searches $\ell_2 \in \{1,10,100\}$; histogram gradient boosting (\texttt{HistGradientBoostingClassifier}) searches $(15\text{ leaves}, \ell_2=1, 120\text{ trees})$ and $(31\text{ leaves}, \ell_2=10, 180\text{ trees})$. Model selection maximizes pre-window smoothed log gain. The selected logit is refitted on the full pool, while the boosted model is refitted on the first 90\% and temperature-calibrated on the final 10\% block. Table~\ref{tab:refs} reports conditional logit for F18 and the equal-weight logit-plus-boosted score ensemble (before within-race softmax) for FAGENT and FV5.

\paragraph{External probability references.}
The calibrated expert forecast transforms previous-evening expert odds (JRDB source KYI) via a power exponent fitted on historical races. Pre-race market probabilities normalize inverse quotes from the last pre-off snapshot (TYB), and closing market probabilities normalize final official odds.

\paragraph{Software and numerical implementation.}
Offline baselines are implemented in Python with SciPy (L-BFGS-B) and scikit-learn; exact package versions, solver tolerances, and random seeds are provided in the released repository.

\subsection{Scoring, costs, settlement, and inference}
\label{app:evaluation-details}

\subsubsection{Data tools and tariff}
\label{app:tariff}
Nine data tools expose cards, recent form, paginated race and horse histories, workouts, aggregate statistics, coverage, connections, and released results. One \emph{evidence cell} is a delivered scalar slot (including nulls and billed IDs, excluding duplicate metadata); a race card with $n$ runners contains $25+43n$ cells. Within a trajectory, repeat queries with identical tool, canonical arguments, and response digest incur zero additional cell fees or cell allowance consumption, though tool-call fees still apply. Base rates are $0.01$ score-yen per cell, $0.5$ per data-tool call, and $0.2$ per sandbox CPU-second, scaled by a common $\times 0.5$ tariff multiplier (control tools are free).

\paragraph{LLM-as-policy accounting.}
\label{app:acquisition-tariff}
In LLM-as-policy, marginal cell charges depend on the native acquisition path and the trajectory's owned cache: expanding recent-history depth ($1 \to 3 \to 5$ starts) or computing condition statistics charges only for previously unseen records and cells, while distinct native operations (e.g., race card vs.\ whole-race workout purchase) retain their respective call and cell fees. Billable CPU time covers generated-policy execution only; updater calls and synthetic validations incur no score-yen fees.

\subsubsection{Common workflow rescoring}
\label{app:acquisition-scoreyen}
To compare code-as-policy and LLM-as-policy on a shared utility scale, saved predictions from both workflows are rescored without the code-as-policy delivery gate $I_r$:
\begin{align*}
\mathrm{Gross}_{\mathrm{cmp}} &= 100\sum_r \left[g_r+h_r-\frac{1}{n_r}\right], \\
\mathrm{Net}_{\mathrm{cmp}} &= \mathrm{Gross}_{\mathrm{cmp}}-C,
\end{align*}
where $C$ applies the common $\times 0.5$ tariff (including one-time warmup costs) under each workflow's cache accounting rules (Appendix~\ref{app:tariff}).

\subsubsection{Flat-WIN settlement}
\label{app:flat-settlement}
The code-as-policy flat-WIN diagnostic (Table~\ref{tab:money}) stakes JPY 100 on the first surviving ranked starter per race and settles at official JRA dividends (including takeout), pooling stakes and payouts across each model's matched trajectories before computing ROI.

\subsubsection{Capped-Kelly settlement}
\label{app:kelly}
In capped-Kelly settlement (Figure~\ref{fig:teaser}), each trajectory starts with a JPY 100,000 bankroll. On day $d$ with start-of-day bankroll $B_d$, the stake on a race's sealed top pick is $B_d\min\{f_r,0.01\}$ where $f_r=\max\{0,(q_rO_r-1)/(O_r-1)\}$, $q_r$ is the sealed win probability, and $O_r$ is the decimal quote from the last pre-race (TYB) snapshot. Stakes are floored to JPY 100 tickets, capped by available cash, refunded on scratches, and settled at official WIN dividends at day close.

\subsubsection{Workflow comparison under shared betting rules}
\label{app:workflow-settlement}
In the nine-model workflow comparison (Section~\ref{sec:workflow-betting}), both workflows are settled on their original sealed top picks without scratch promotion (withdrawals are refunded) and using raw saved probabilities $q_r$ paired with archived TYB quotes $O_r$ (skipping races with missing quotes). Each trajectory is simulated in a separate account before averaging terminal P\&L. Turnover ROI divides total profit by total original stakes (including refunded stakes). For risk metrics, per-race profits are first averaged equally across a configuration's $n$ trajectories: maximum drawdown (MDD) is the largest drop from a running peak of the cumulative mean profit path (starting at zero), and the annualized P\&L Sharpe ratio is
\[
\mathrm{SR}_{\text{P\&L}}=\frac{\mu}{s}\sqrt{3454},
\]
where $\mu$ and $s$ are the mean and sample SD of the averaged per-race JPY profit series scaled by the nominal JRA annual race count (3,454 races/year), which assumes i.i.d.\ per-race returns~\citep{sharpe1994ratio,lo2002statistics}.

\subsubsection{Paired statistical inference}
\label{app:stats}
For paired \Live{}$-$\Frozen{} contrasts in Table~\ref{tab:main}, let $d_j$ be the mean difference on day $j$ (after averaging repeats within each race) across $n_j$ unique races, with $N=\sum_{j=1}^g n_j=324$ and $g=10$ days. The point estimate and day-clustered standard error~\citep{cameron2015practitioner} are
\begin{align*}
\hat\Delta&=\frac{\sum_j n_jd_j}{N},\\
\widehat{\mathrm{SE}}&=\frac{\sqrt{\frac{g}{g-1}\sum_j[n_j(d_j-\hat\Delta)]^2}}{N}.
\end{align*}
Table~\ref{tab:main} reports $\hat\Delta\pm\widehat{\mathrm{SE}}$, and two-sided $t$-test $p$-values are adjusted across the \nModels{} configurations via Holm's step-down procedure~\citep{holm1979simple} (zero-variance contrasts with $\hat\Delta=0$ receive $p=1$).

\section{Extended Results and Diagnostics}
\label{app:extended-results}
\label{app:acquisition-diagnostics}
The analyses below supplement Section~\ref{sec:experiments}; each subsection states the sample it uses.

\subsection{Reference scaling and workflow forecast quality}
\label{app:diagnostics}

\paragraph{Reference scaling.}
Table~\ref{tab:ceiling} evaluates offline references on the 324-race main window across training-set sizes ($N \in [100, 20{,}730]$) and fitting methods. With FV5 features, HGB trails conditional logit at 3,000 races, scoring 0.571 versus 0.577 nats, and first leads at 10,000 races, scoring 0.606 versus 0.584.
\begin{table}[!tbp]
\caption{Smoothed log gain on the 324-race main window by training-set size and feature representation. F18: 18 hand-built features; FAGENT: agent-inspired features chosen post hoc; FV5: broad factual features (definitions in Appendix~\ref{app:refs}). Logit: conditional logit; HGB: histogram gradient boosting; $\oplus$: logit averaging before softmax. Training subsets differ in size and historical composition; columns vary feature representation and fitting method.}
\label{tab:ceiling}
\centering
\renewcommand{\arraystretch}{1.14}
\small
\setlength{\tabcolsep}{3pt}
\begin{tabular*}{\textwidth}{@{\extracolsep{\fill}}lcccccc@{}}
\toprule
& \multicolumn{1}{c}{\textbf{F18}} & \multicolumn{2}{c}{\textbf{FAGENT}} & \multicolumn{3}{c}{\textbf{FV5}} \\
\cmidrule(lr){2-2}\cmidrule(lr){3-4}\cmidrule(lr){5-7}
Train $N$ & Logit & Logit & Logit$\oplus$HGB & Logit & HGB & Logit$\oplus$HGB \\
\midrule
100 & \cellcolor[HTML]{E7EFEB}0.411 & \cellcolor[HTML]{DBE8E2}0.470 & \cellcolor[HTML]{E1ECE7}0.438 & \cellcolor[HTML]{EAF2EE}0.394 & \cellcolor[HTML]{F8FAF9}0.327 & \cellcolor[HTML]{E3EDE8}0.431 \\
300 & \cellcolor[HTML]{E0ECE6}0.442 & \cellcolor[HTML]{D7E6DF}0.487 & \cellcolor[HTML]{DCE9E3}0.462 & \cellcolor[HTML]{D6E5DE}0.492 & \cellcolor[HTML]{E8F0EC}0.405 & \cellcolor[HTML]{D6E5DE}0.493 \\
1,000 & \cellcolor[HTML]{DCE9E3}0.461 & \cellcolor[HTML]{D4E4DC}0.501 & \cellcolor[HTML]{D6E5DE}0.494 & \cellcolor[HTML]{CDE0D7}0.536 & \cellcolor[HTML]{DBE9E2}0.466 & \cellcolor[HTML]{CDE0D7}0.535 \\
3,000 & \cellcolor[HTML]{DCE9E3}0.462 & \cellcolor[HTML]{D4E4DC}0.502 & \cellcolor[HTML]{D2E3DB}0.510 & \cellcolor[HTML]{C4DBD0}0.577 & \cellcolor[HTML]{C6DBD1}0.571 & \cellcolor[HTML]{C0D8CD}0.597 \\
10,000 & \cellcolor[HTML]{DCE9E3}0.463 & \cellcolor[HTML]{D4E4DC}0.503 & \cellcolor[HTML]{D0E2D9}0.519 & \cellcolor[HTML]{C3DACF}0.584 & \cellcolor[HTML]{BED7CB}0.606 & \cellcolor[HTML]{BDD6CA}0.613 \\
all (20,730) & \cellcolor[HTML]{DCE9E3}0.462 & \cellcolor[HTML]{D4E4DC}0.503 & \cellcolor[HTML]{D0E1D9}0.523 & \cellcolor[HTML]{C3DACF}0.584 & \cellcolor[HTML]{C0D8CC}0.598 & \cellcolor[HTML]{BED7CB}0.608 \\
\bottomrule
\end{tabular*}

\end{table}

\subsubsection{Forecast quality across workflows}
\label{app:workflow-forecast-quality}

\paragraph{Scoring and delivery breakdown.}
Table~\ref{tab:workflow-forecast-quality} reports mean smoothed log gain, Top-1 win rate, and Top-3 precision ($\pm$ between-run sample SDs) across the three trajectories per model and workflow. Across all 8,748 LLM-as-policy forecasts ($9 \times 3 \times 324$), native delivery tags record 8,453 direct forecasts ($96.63\%$; ranging from $88.68\%$ on GLM-5.3 to $99.79\%$ on GPT-6.1-Sol), 159 repaired forecasts, 96 execution fallbacks, and 40 uniform submissions. Across the matched 8,748 code-as-policy forecasts, 8,692 are \texttt{predictor} deliveries ($99.36\%$) and 56 are \texttt{silent\_uniform} outputs. Delivery does not explain the quality ranking: GLM-5.3-Flash delivers 98.66\% of its forecasts directly, close to GPT-6.1-Sol's 99.79\%, yet scores 0.258 against 0.502 nats, and the lowest direct share (GLM-5.3, 88.68\%) does not coincide with the lowest log gain (GPT-6-Luna, 0.112 nats).

\begin{table}[t]
\centering
\small
\setlength{\tabcolsep}{4pt}
\renewcommand{\arraystretch}{1.08}

\caption{
Forecast quality across nine models under two workflows.
Results are reported as mean $\pm$ sample SD over
$n=3$ independent runs on the same \nRaces{} races.
Bold indicates the higher mean within each model and metric,
without implying statistical significance.
}
\label{tab:workflow-forecast-quality}

\begin{tabular*}{\textwidth}
{@{\extracolsep{\fill}}llccc@{}}
\toprule
Model & Workflow & Log gain (nats)
& Top-1 (\%) & Top-3 precision (\%) \\
\midrule

\multirow{2}{*}{\textbf{DeepSeek-V4-Flash}}
&  \CodePolicyWorkflow
& $0.269$\,{\scriptsize$\pm$0.180}
& $\mathbf{22.94}$\,{\scriptsize$\pm$2.28}
& $41.60$\,{\scriptsize$\pm$3.08} \\
& \LLMPolicyWorkflow
& $\mathbf{0.297}$\,{\scriptsize$\pm$0.030}
& $22.22$\,{\scriptsize$\pm$1.63}
& $\mathbf{41.70}$\,{\scriptsize$\pm$0.97} \\
\addlinespace[4pt]

\multirow{2}{*}{\textbf{DeepSeek-V4.1-Flash}}
&  \CodePolicyWorkflow
& $0.233$\,{\scriptsize$\pm$0.052}
& $21.09$\,{\scriptsize$\pm$4.03}
& $37.93$\,{\scriptsize$\pm$1.70} \\
& \LLMPolicyWorkflow
& $\mathbf{0.277}$\,{\scriptsize$\pm$0.015}
& $\mathbf{21.50}$\,{\scriptsize$\pm$2.62}
& $\mathbf{41.87}$\,{\scriptsize$\pm$0.45} \\
\addlinespace[4pt]

\multirow{2}{*}{\textbf{GLM-5.3}}
&  \CodePolicyWorkflow
& $\mathbf{0.422}$\,{\scriptsize$\pm$0.065}
& $\mathbf{25.82}$\,{\scriptsize$\pm$0.99}
& $\mathbf{44.03}$\,{\scriptsize$\pm$1.28} \\
& \LLMPolicyWorkflow
& $0.263$\,{\scriptsize$\pm$0.034}
& $21.19$\,{\scriptsize$\pm$3.40}
& $40.12$\,{\scriptsize$\pm$1.43} \\
\addlinespace[4pt]

\multirow{2}{*}{\textbf{GLM-5.3-Flash}}
&  \CodePolicyWorkflow
& $\mathbf{0.410}$\,{\scriptsize$\pm$0.015}
& $\mathbf{25.00}$\,{\scriptsize$\pm$0.82}
& $\mathbf{44.14}$\,{\scriptsize$\pm$0.00} \\
& \LLMPolicyWorkflow
& $0.258$\,{\scriptsize$\pm$0.012}
& $21.09$\,{\scriptsize$\pm$1.78}
& $40.57$\,{\scriptsize$\pm$0.86} \\
\addlinespace[4pt]

\multirow{2}{*}{\textbf{GPT-6-Astra}}
&  \CodePolicyWorkflow
& $\mathbf{0.520}$\,{\scriptsize$\pm$0.009}
& $\mathbf{30.14}$\,{\scriptsize$\pm$0.47}
& $\mathbf{48.15}$\,{\scriptsize$\pm$1.16} \\
& \LLMPolicyWorkflow
& $0.491$\,{\scriptsize$\pm$0.028}
& $26.65$\,{\scriptsize$\pm$1.55}
& $46.16$\,{\scriptsize$\pm$0.49} \\
\addlinespace[4pt]

\multirow{2}{*}{\textbf{GPT-6-Luna}}
&  \CodePolicyWorkflow
& $\mathbf{0.236}$\,{\scriptsize$\pm$0.165}
& $\mathbf{20.58}$\,{\scriptsize$\pm$4.50}
& $\mathbf{40.12}$\,{\scriptsize$\pm$3.12} \\
& \LLMPolicyWorkflow
& $0.112$\,{\scriptsize$\pm$0.146}
& $12.96$\,{\scriptsize$\pm$5.56}
& $32.03$\,{\scriptsize$\pm$7.08} \\
\addlinespace[4pt]

\multirow{2}{*}{\textbf{GPT-6-Sol}}
&  \CodePolicyWorkflow
& $\mathbf{0.474}$\,{\scriptsize$\pm$0.007}
& $\mathbf{28.19}$\,{\scriptsize$\pm$1.17}
& $\mathbf{45.68}$\,{\scriptsize$\pm$1.29} \\
& \LLMPolicyWorkflow
& $0.367$\,{\scriptsize$\pm$0.027}
& $21.71$\,{\scriptsize$\pm$1.70}
& $43.21$\,{\scriptsize$\pm$0.36} \\
\addlinespace[4pt]

\multirow{2}{*}{\textbf{GPT-6.1-Sol}}
&  \CodePolicyWorkflow
& $\mathbf{0.524}$\,{\scriptsize$\pm$0.003}
& $\mathbf{28.40}$\,{\scriptsize$\pm$1.11}
& $\mathbf{47.87}$\,{\scriptsize$\pm$0.24} \\
& \LLMPolicyWorkflow
& $0.502$\,{\scriptsize$\pm$0.008}
& $27.06$\,{\scriptsize$\pm$0.64}
& $45.30$\,{\scriptsize$\pm$0.59} \\
\addlinespace[4pt]

\multirow{2}{*}{\textbf{MiMo-V2.6-Pro}}
&  \CodePolicyWorkflow
& $\mathbf{0.438}$\,{\scriptsize$\pm$0.058}
& $\mathbf{26.95}$\,{\scriptsize$\pm$0.64}
& $\mathbf{45.64}$\,{\scriptsize$\pm$1.21} \\
& \LLMPolicyWorkflow
& $0.281$\,{\scriptsize$\pm$0.040}
& $23.05$\,{\scriptsize$\pm$2.87}
& $41.53$\,{\scriptsize$\pm$1.93} \\

\bottomrule
\end{tabular*}
\end{table}

\subsection{Revision and frozen controls}
\label{app:revision-results}
\paragraph{Predictor revision.}
Table~\ref{tab:main} reports mean \Live{} log gain and the paired \Live{}--\Frozen{} difference across all \nARuns{} code-as-policy pairs. Figure~\ref{fig:code-policy-forecast-quality} displays the five-run spread and paired means for every configuration. The leading configurations have tightly clustered runs and nearly coincident \Live{} and \Frozen{} means; several lower-ranked configurations show wider spreads and larger mean shifts. All \Live{} runs remain below the closing-market reference on this evaluation window. Paired significance tests use the day-clustered procedure in Appendix~\ref{app:evaluation-details}; qualitative code-revision traces appear in Appendix~\ref{app:case-studies}.

\paragraph{Biased warmup samples (code-as-policy).}
The support history spans 2015-01-04 to 2023-12-28, and a runner's visible form contains only starts inside the archive. None of the runners in the oldest 50 races has a visible past start, only 10\% of those in the oldest 220 races do, and coverage reaches 85--97\% from April 2015 onward (85.7\% in the newest 150 support races, 87.9\% in the main window). A warmup that pages the history oldest-first therefore trains on races where form is almost absent. Across the 130 \Frozen{} arms of the earlier 26-configuration cohort, ten warmups drew most of their form-bearing training races from this edge, and eight of them score below the median of the same model's other rounds (median difference $-0.142$ nats). In the extreme case, GLM-5.3-Flash round 5 fits a 20-feature logit on the 220 oldest races, whose mean number of visible starts is 0.10; in the window it favors debutants and scores $-1.436$ nats, with a mean top-pick probability of 0.63. This is a left-truncation effect of the archive rather than ordinary overfitting. A future protocol should start the support set later or report form coverage to the agent.

\subsubsection{Matched controls for evidence-plan revision (LLM-as-policy)}
\label{app:matched-frozen-acquisition}
\label{app:discussion}

\paragraph{Protocol and reward--cost trade-offs.}
We compare five matched pairs: DeepSeek-V4-Flash R1, R2, R4, and R6,
and GPT-6-Astra R1. Within each pair, both arms use the same
forecasting model, initial evidence plan $P_0$, and 324-race
schedule. The Frozen-$P_0$ arm uses $P_0$ throughout, while the
revisable arm can update its evidence plan between race days.
The two arms generate forecasts separately. The Frozen arm reuses
$P_0$ without repeating warmup, and the score-yen accounting includes
the same historical warmup cost once per arm. The ledger records exactly the data and call fees attributable to each plan's queries; total fees also include metered CPU time for plan execution. Differences in gross reward and log gain also include run-to-run forecasting variation, because both arms forecast each race with separate LLM calls; unlike the code-as-policy \Frozen{} arm, which re-executes the same predictor deterministically, these pairs do not separate that variation from the effect of revision.

Under the common tariff, the revisable arm had lower weighted research
fees and a lower gross score in four of the five pairs. Its net score was higher
in all four DeepSeek-V4-Flash pairs and lower in GPT-6-Astra R1.
The revisable arm had higher mean smoothed log gain in two pairs and
higher Top-3 precision in four. Its retrospective P\&L was higher in
three pairs under flat WIN and two under capped Kelly. Fee savings came from reducing \texttt{condition\_stats} queries or shortening \texttt{recent\_history} depth (e.g., saving $4{,}411.96$ score-yen in DeepSeek-V4-Flash R4).

\subsubsection{Evidence acquisition across LLM-as-policy models}
\label{app:acquisition-by-model}

We analyze the 27 LLM-as-policy runs in the balanced workflow
cohort (Appendix~\ref{app:models}), with three complete 324-race runs
per model. Table~\ref{tab:acquisition-by-model} orders the models by
mean smoothed log gain across their three runs.

\begin{table}[t]
\centering
\small
\setlength{\tabcolsep}{4pt}
\caption{Evidence acquisition by model across the 27 LLM-as-policy
runs. Recent-history coverage is the percentage of runner
entries with an explicit history query. Depth is the mean final
request depth among those entries. The three rightmost percentages
use all races as the denominator. Paid history denotes a separate
paid history retrieval triggered by a condition-statistics query.}
\label{tab:acquisition-by-model}
\begin{tabular*}{\textwidth}{@{\extracolsep{\fill}}lrrrrrr@{}}
\toprule
 & & \multicolumn{2}{c}{Recent history} & Workout data
   & \multicolumn{2}{c}{Condition statistics} \\
\cmidrule(lr){3-4}\cmidrule(lr){5-5}\cmidrule(l){6-7}
Model & Log gain & Coverage (\%) & Depth & Purchased (\%)
      & Queried (\%) & Paid history (\%) \\
\midrule
GPT-6.1-Sol         & 0.502 & 100.0 & 3.41 & 100.0 & 81.9 &  0.0 \\
GPT-6-Astra         & 0.491 & 100.0 & 2.42 & 100.0 & 38.7 &  0.0 \\
GPT-6-Sol           & 0.367 & 100.0 & 1.25 & 100.0 &  0.0 &  0.0 \\
DeepSeek-V4-Flash   & 0.297 & 100.0 & 2.95 & 100.0 & 67.0 & 15.6 \\
MiMo-V2.6-Pro       & 0.281 &  96.4 & 3.77 & 100.0 & 76.4 & 10.3 \\
DeepSeek-V4.1-Flash & 0.277 & 100.0 & 2.01 &  79.0 & 34.8 &  8.3 \\
GLM-5.3             & 0.263 &  99.9 & 3.30 &  96.7 & 77.7 & 32.7 \\
GLM-5.3-Flash       & 0.258 & 100.0 & 1.33 & 100.0 & 39.0 & 31.1 \\
GPT-6-Luna          & 0.112 &  57.7 & 1.07 &  27.2 &  0.0 &  0.0 \\
\bottomrule
\end{tabular*}
\end{table}

\begin{table}[t]
\centering
\small
\setlength{\tabcolsep}{2pt}
\caption{Evidence-plan changes per model in the 27 LLM-as-policy runs. A change is a day (2--10) whose active plan differs in content from the previous day's. \emph{Changes} is the mean number of changes over three runs; \emph{Last change} gives the range of the final change day.}
\label{tab:plan-cadence}
\begin{tabular*}{\columnwidth}{@{\extracolsep{\fill}}lrrc@{}}
\toprule
Model & Log gain & Changes & Last change \\
\midrule
GPT-6.1-Sol         & 0.502 & 1.0 & 2     \\
GPT-6-Astra         & 0.491 & 1.0 & 2     \\
GPT-6-Sol           & 0.367 & 6.0 & 4--10 \\
DeepSeek-V4-Flash   & 0.297 & 6.3 & 10    \\
MiMo-V2.6-Pro       & 0.281 & 4.3 & 6--10 \\
DeepSeek-V4.1-Flash & 0.277 & 7.0 & 7--10 \\
GLM-5.3             & 0.263 & 4.7 & 6--10 \\
GLM-5.3-Flash       & 0.258 & 4.7 & 4--10 \\
GPT-6-Luna          & 0.112 & 5.7 & 9     \\
\bottomrule
\end{tabular*}
\end{table}

\paragraph{Recent history and workout data.}
The three models with the highest mean log gain queried recent history
for every horse and purchased workout data for every race. Their
mean final history depths ranged from 1.25 to 3.41 starts per horse.
The bottom three models queried history for 85.9\% of horses and
purchased workout data in 74.6\% of races. Most of the coverage gap
came from GPT-6-Luna. Together, the two GLM models queried recent
history for 99.9\% of horses and purchased workout data in 98.4\%
of races.

\paragraph{Condition statistics and track context.}
The top and bottom groups requested condition statistics in 40.2\%
and 38.9\% of races, respectively. All 7,110 requests in the top group
used previously acquired history. In the bottom group, these requests
triggered additional paid history retrieval in 620 races (21.3\%).
GPT-6-Sol ranked third in mean log gain with zero requests for condition
statistics. All nine models acquired track context in every race at
0.25 score-yen per race. Direct history retrieval and history retrieval
triggered by condition statistics together accounted for 92.9\% of
fees across the four acquisition domains.

\paragraph{Summary of acquisition choices.}
A fixed evidence plan can still vary its queries from race to race.

\subsection{Schema blinding and hindsight controls}
\label{app:priors}
This appendix examines sensitivity to field semantics and audits agent outputs for evidence of result recall (Section~\ref{sec:quality}). The no-warmup and blinded runs use the same \nRaces{} races and replay clock as the main cohort, with three changes. Agents get no labeled warmup races, they are scored by log gain alone with no fees, and each model runs three times. We therefore report these runs separately from Table~\ref{tab:main}.

\paragraph{Blinded schemas.}
In a blinded variant, field names are replaced by opaque identifiers, categorical values by tokens, numeric values by a random affine map with random sign, and token prefixes are stripped. Blinding lowers the no-warmup window mean by $0.011$ (Claude Opus 5), $0.063$ (GPT-6-Astra), $0.072$ (DeepSeek-V4.1-Flash), and $0.135$ nats (GLM-5.3-Flash). Strong models recover key fields from structure at initialization: Opus noticed that past finishing position is a permutation within each past race, and Astra inverted the affine map from the integer lattice. Blinding is therefore partial for integer-valued fields.

\paragraph{Re-identification and output audit.}
Race, horse, jockey, trainer, and pedigree identifiers are HMAC tokens, and race records use relative day indices. Code-as-policy agents, and LLM-as-policy per-race forecast and repair requests, receive no calendar dates. The LLM-as-policy warmup planner receives the start date of the evaluation window, and its post-release updater receives the date and release time of the completed race day. In our re-identification check, matching the visible day index, venue code, and meeting day to the JRA calendar uniquely identified the window start from the two race days examined. Of the 360 races checked, 350 were uniquely identifiable within their day and venue from visible race attributes. Races are therefore re-identifiable in principle by an agent that has memorized the calendar and the results. We scanned every agent-written message, reasoning trace where available, tool argument, and saved predictor across 431 code-as-policy \Live{} trajectories, covering the main cohort and the control experiments, for calendar reconstruction, racecourse and graded-race names, ISO dates, and result recall. No trajectory maps day indices to dates or recalls a result. Racecourse names appear in six trajectories, all in generic discussion of draw bias or code meanings. ISO dates appear in four, all model-invented strings found in no tool response. Because code-as-policy forecasts are sealed by code that runs without an LLM, any use of memorized outcomes would have to be written into a saved predictor, where it can be audited. LLM-as-policy offers no such guarantee; its per-race forecasts have not yet been scanned.

\subsection{Odds-visible control}
\label{app:odds}

\paragraph{Design.}
This preregistered arm tests whether agents can improve on the market price once they see it. It adds the last pre-race odds snapshot to every race card and keeps the main cohort's races, fees, and objective, so agents decide at the snapshot time instead of the previous evening. GPT-6-Astra, Claude Opus 5.5, GLM-5.3-Flash, and DeepSeek-V4-Pro each ran three code-as-policy \Live{} trajectories with their \Frozen{} controls, and DeepSeek-V4.1-Flash ran the same design as a supplementary model. All 30 trajectories delivered a validated predictor on every race. Table~\ref{tab:odds} reports the results.

\paragraph{Forecasts collapse onto the price.}
With odds visible, every model's smoothed log gain rises to between 0.701 and 0.713, next to the snapshot market's 0.700, and the spread across models falls from 0.30 to 0.01 nats. No \Live{} program reads race form, and most keep the market's ordering of runners. Four of the five models beat the raw snapshot price by 0.003 to 0.013 nats, but the margin is a calibration of that price. A one-parameter recalibration that we fit on 2023 races gains 0.010 nats by itself, and no model significantly beats the recalibrated price; we added this comparison after the preregistered analysis. All models remain 0.04 to 0.06 nats below the closing market. Once odds are visible, the benchmark therefore stops separating models, and the remaining differences measure how an agent reshapes the price rather than how it reads the evidence.

\paragraph{Market-level forecasts do not make money.}
With odds visible, the top pick is the snapshot favorite in 98 to 100 percent of races, so a flat 100-yen bet on the top pick returns between $-22.2\%$ and $-24.7\%$ at official dividends. These returns match the favorite's own $-22.9\%$ and sit close to the 20\% takeout. GPT-6-Astra and Claude Opus 5.5 forecast far better with odds, yet their flat returns fall from $-5.0\%$ and $-10.5\%$ to $-22.9\%$. Value bets almost disappear. Without odds, nearly every race contains a runner whose forecast clears the break-even price, and these 2,200 to 2,500 bets per run lose 40 to 48 percent at official dividends. With odds visible, three of the five models place no such bet. Even at closing odds with the takeout removed, full Kelly betting on these forecasts would lose about 0.05 nats per race for every model.

\paragraph{Scope.}
Under the benchmark's fees, agents that see the price stop buying other evidence, so this arm does not test whether free evidence would let agents add information beyond the price. A zero-fee arm on the same races, which uses a richer card and a profit objective, comes closest at 0.011 nats above the recalibrated price, and this margin is not significant.

\begin{table}[t]
\caption{\textbf{With odds visible, the models converge on the market.} Each row averages three code-as-policy \Live{} runs on the same \nRaces{} races, and \emph{No odds} uses the same model's first three main-cohort runs. \emph{Top-pick odds} is the median snapshot odds of each race's top pick. $^\dagger$Supplementary model outside the preregistered family.}
\label{tab:odds}
\centering
\footnotesize
\setlength{\tabcolsep}{4pt}
\begin{tabular}{@{}lcccc@{}}
\toprule
 & \multicolumn{2}{c}{Log gain} & \multicolumn{2}{c}{Top-pick odds} \\
\cmidrule(lr){2-3}\cmidrule(l){4-5}
Model & No odds & Odds & No odds & Odds \\
\midrule
GPT-6-Astra & 0.520 & 0.703 & 3.3 & 2.6 \\
Claude Opus 5.5 & 0.467 & 0.712 & 3.5 & 2.6 \\
GLM-5.3-Flash & 0.410 & 0.710 & 3.6 & 2.6 \\
DeepSeek-V4-Pro & 0.223 & 0.701 & 4.5 & 2.6 \\
DeepSeek-V4.1-Flash$^\dagger$ & 0.233 & 0.713 & 4.9 & 2.6 \\
\midrule
Range across models & 0.297 & 0.012 & 1.6 & 0.0 \\
Snapshot market & -- & 0.700 & -- & 2.6 \\
\bottomrule
\end{tabular}
\end{table}

\subsection{Retrospective returns and risk}
\label{app:return-results}
\paragraph{Code-as-policy flat-WIN and capped-Kelly diagnostics.}
\begin{table}[!t]
\centering
\footnotesize
\setlength{\tabcolsep}{4pt}
\renewcommand{\arraystretch}{1.14}
\caption{Retrospective settlement at official dividends (takeout included) for flat 100-yen WIN or PLACE tickets on the top-ranked starter in each of 324 races. The table covers \nTakeoutRuns{} code-as-policy \Live{} trajectories from \nTakeoutModels{} models. The first 16 models use three rounds from the 23 September archive. The final four use all five matched main-cohort rounds. Pooled WIN ROI across all tickets is $\allWinROI\%$.}
\label{tab:money}

\begin{tabular*}{\columnwidth}{@{\extracolsep{\fill}}lcc@{}}
\toprule
Model & Win ROI (\%) & Place ROI (\%) \\
\midrule

\multicolumn{3}{l}{\textbf{Three rounds ($n=3$)}} \\

GPT-6-Astra & $-$5.0 & $-$16.3 \\
Claude Fable 5.1 & $-$12.7 & $-$15.6 \\
GPT-5.6-Sol & $-$7.6 & $-$14.9 \\
GPT-6-Sol & $-$8.1 & $-$15.9 \\
Claude Opus 5 & $-$10.0 & $-$14.1 \\
Claude Opus 5.5 & $-$10.5 & $-$15.9 \\
GLM-5.3 & $-$17.0 & $-$20.6 \\
GLM-5.3-Flash & $-$18.7 & $-$21.7 \\
Kimi-K3 & $-$26.3 & $-$24.0 \\
GPT-5.6-Luna & $-$16.6 & $-$21.3 \\
DeepSeek-V4-Flash & $-$17.9 & $-$21.0 \\
GPT-6-Luna & $-$24.8 & $-$21.2 \\
DeepSeek-V4-Pro & $-$29.3 & $-$29.2 \\
MiMo-V2.5-Pro & $-$22.3 & $-$15.8 \\
MiMo-V2.6-Flash & $-$30.4 & $-$21.3 \\
GPT-5.6-Terra & $-$37.7 & $-$19.8 \\

\addlinespace[4pt]
\multicolumn{3}{l}{\textbf{Five rounds ($n=5$)}} \\

Gemma-4-26B-A4B & $-$27.2 & $-$24.9 \\
Nemotron-3.5-Lightning & $-$25.0 & $-$44.9 \\
Qwen3.5-9B & $-$20.4 & $-$29.3 \\
Qwen3.5-35B-A3B & $-$40.7 & $-$31.0 \\

\midrule

Favourite at pre-off snapshot & $-$22.9 & $-$16.9 \\
Favourite at closing & $-$19.4 & $-$17.4 \\
Uniform random pick & $-$38.2 & $-$27.9 \\

\bottomrule
\end{tabular*}
\end{table}

\begin{figure}[!t]
\centering\includegraphics[width=\textwidth]{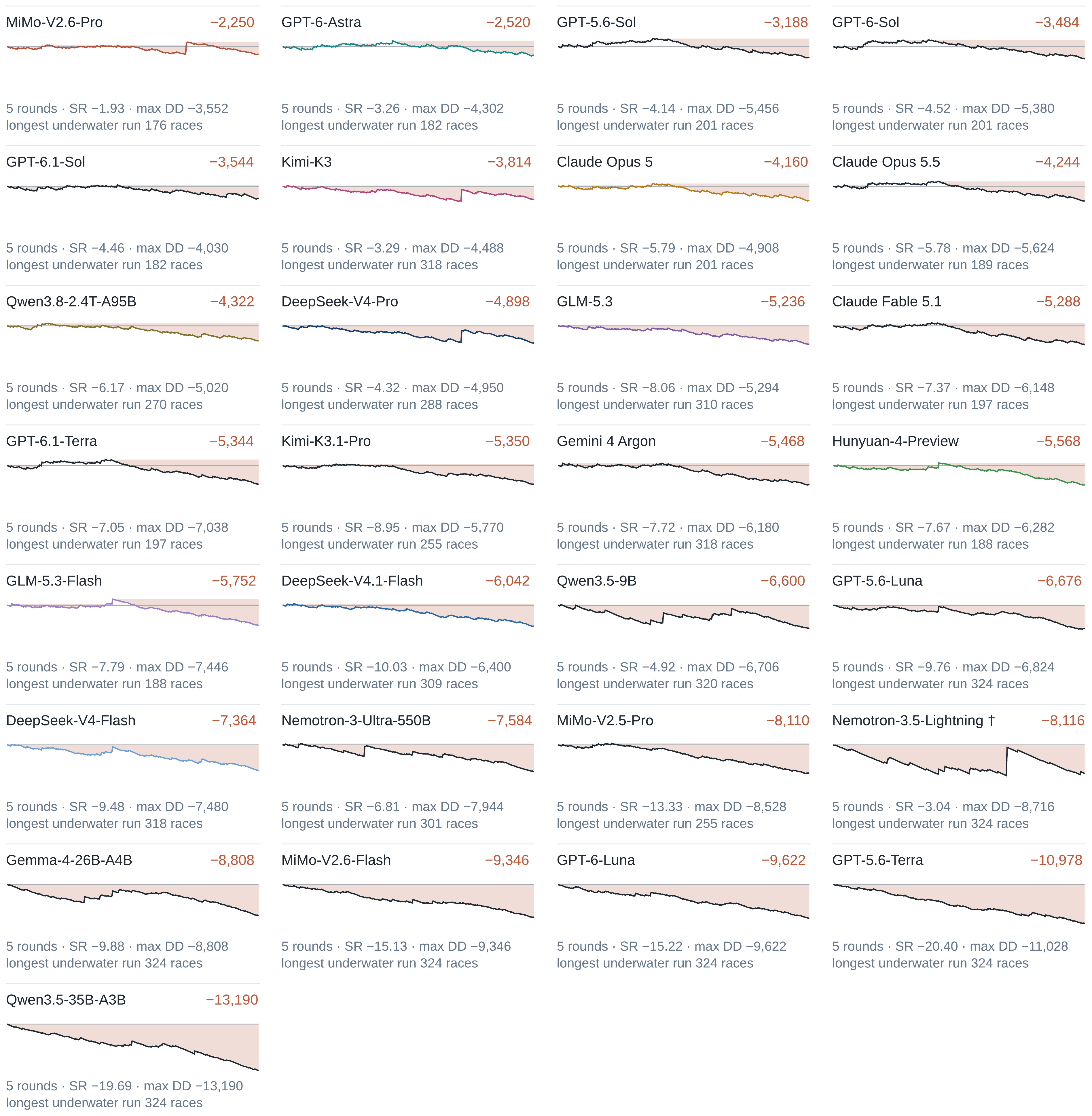}
\caption{Mean cumulative profit across five repeats from flat JPY 100 WIN stakes at official dividends (\nModels{} configurations, \nARuns{} runs). Shading indicates drawdown from the running peak; footers report repeat counts, annualized P\&L Sharpe ratio (SR), maximum drawdown (JPY), and longest underwater run. \textdagger{} marks Nemotron-3.5-Lightning, which delivered only uniform fallback forecasts.}
\label{fig:equity}
\vspace{-5pt}
\end{figure}

Table~\ref{tab:money} reports WIN and PLACE ROI for \nTakeoutRuns{} trajectories from \nTakeoutModels{} models (pooled WIN ROI $\allWinROI\%$). Figure~\ref{fig:equity} tracks mean cumulative flat-WIN profit across all \nModels{} configurations (\nARuns{} runs): every configuration finishes below zero, and every annualized P\&L Sharpe ratio is negative, ranging from $-1.93$ (MiMo-V2.6-Pro) to $-20.40$ (GPT-5.6-Terra; backing the closing favorite yields $-9.42$). Even the least negative configurations spend most of the window in drawdown (e.g., 182 of 324 races for GPT-6-Astra). Under 1\%-capped Kelly (Figure~\ref{fig:teaser}), the 1\% cap binds on 96\% of placed bets, so the rule effectively stakes 1\% of bankroll on the top pick whenever $q_r O_r > 1$ (an average of 163 bets per 324-race trajectory). No trajectory goes bankrupt, 13 of \nARuns{} individual runs finish in profit, and all \nModels{} configuration means are negative.

\begin{figure}[t]
\centering
\includegraphics[width=\textwidth]{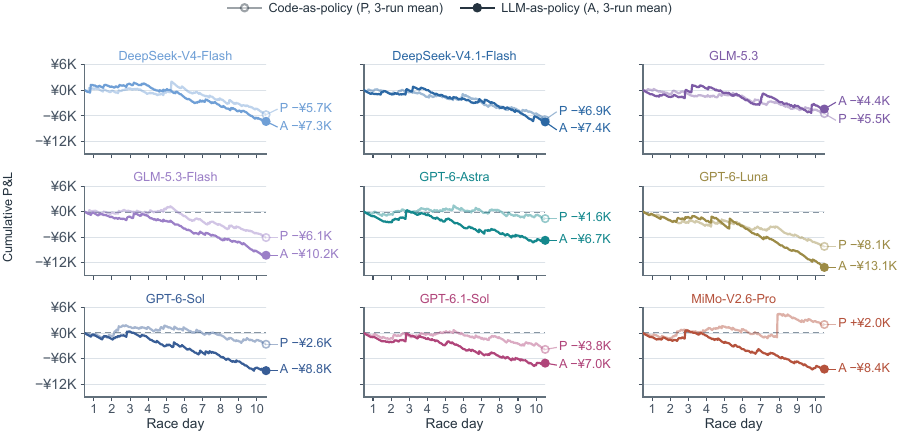}
\caption{Mean cumulative flat-WIN P\&L (JPY 100 stake per race, withdrawals refunded) across the balanced nine-model workflow cohort ($n=3$, rounds 1--3; dashed \texttt{P}: code-as-policy, solid \texttt{A}: LLM-as-policy).}
\label{fig:workflow-flat-profit}
\end{figure}

\paragraph{Workflow returns and risk.}
Unless noted otherwise, workflow analyses below use the balanced cohort of nine models, each with three trajectories per workflow (27 per workflow); Appendix~\ref{app:deepseek-return-concentration} alone uses all six completed DeepSeek-V4-Flash LLM-as-policy repeats.
Table~\ref{tab:workflow-betting-risk} reports individual-trajectory P\&L (R1--R3), mean terminal P\&L, pooled turnover ROI, and mean-path risk (MDD and annualized P\&L Sharpe; definitions in Appendix~\ref{app:workflow-settlement}) across the nine shared models. Figure~\ref{fig:workflow-flat-profit} plots the corresponding mean cumulative flat-WIN profit paths.

\subsubsection{Calibration and bet selection (both workflows)}
\label{app:bet-selection}

\begin{samepage}
\paragraph{Selection-induced overconfidence.}
Comparing raw saved probabilities $q_r$ on all non-refunded top picks against the subset selected by the Kelly rule ($q_r O_r > 1$) explains why positive log gains fail to yield positive betting returns. Across all top picks, LLM-as-policy underpredicts observed win rates on seven of nine models (only GPT-6-Luna and GPT-6-Sol overpredict overall). Within the Kelly-selected subset, however, \textbf{both workflows substantially overpredict observed win frequency across all nine models}. For example, GLM-5.3-Flash under LLM-as-policy assigns a mean probability of $16.78\%$ to its Kelly selections, whereas their observed win rate is only $5.21\%$.

\end{samepage}

\paragraph{Longshot exposure and equal-stake check.}
On GLM-5.3-Flash, runners quoted at odds $O_r \ge 10$ account for $59.64\%$ of LLM-as-policy's Kelly turnover versus $27.60\%$ for code-as-policy. Reweighting the 425 recorded LLM-as-policy Kelly bets (422 non-refunded) to flat JPY 100 stakes yields an ROI of $-55.08\%$, virtually identical to its $-56.23\%$ ROI under variable Kelly sizing. The loss therefore persists when all selected tickets receive equal stakes, so it does not come from stake sizing.

\subsubsection{Return concentration across DeepSeek-V4-Flash LLM-as-policy repeats}
\label{app:deepseek-return-concentration}

Across all six completed DeepSeek-V4-Flash LLM-as-policy repeats (R1--R4, R6, R7; R1--R3 belong to the balanced workflow cohort), only R1 finishes with a positive Kelly profit ($+\text{JPY } 1{,}470$, from 13 wins across 144 non-refunded bets), whereas the other five repeats lose JPY $47{,}568$ on average. R1's three largest winning tickets accounted for $55.35\%$ of its
gross payouts. Bets at quoted odds $O_r \ge 10$ earned R1
JPY~$32{,}700$ in net profit, offsetting JPY~$31{,}230$ in losses at
lower odds. The other five repeats lost an average of JPY~$32{,}050$
per run on bets at quoted odds of 10 or higher. Crucially, R1's windfall does not reflect superior probability calibration: its Kelly selections still overpredict wins ($17.88\%$ mean predicted probability vs.\ $9.03\%$ observed), and repeats R2, R3, R4, and R6 achieve lower (better) full-vector Brier scores than R1 despite losing money under Kelly.

\begin{table}[!tp]
\centering
\scriptsize
\setlength{\tabcolsep}{2pt}
\renewcommand{\arraystretch}{1.08}
\caption{Individual-trajectory betting P\&L (R1--R3, in JPY) and aggregate return and mean-path risk across workflows ($n=3$). Bold R1--R3 values mark positive profits; bold aggregate values mark the more favorable workflow within each model and staking rule.}
\label{tab:workflow-betting-risk}

\begin{tabular*}{\textwidth}
{@{\extracolsep{\fill}}llccccccc@{}}
\toprule
& & \multicolumn{3}{c}{Run P\&L (JPY)}
& \multicolumn{2}{c}{Return}
& \multicolumn{2}{c}{Mean-path risk} \\
\cmidrule(lr){3-5}
\cmidrule(lr){6-7}
\cmidrule(lr){8-9}
Model & Workflow
& R1 & R2 & R3
& Mean P\&L & ROI (\%)
& Max.\ DD & P\&L SR \\
\midrule

\multicolumn{9}{l}{\textbf{Flat JPY 100 WIN}} \\

\multirow{2}{*}{DeepSeek-V4-Flash}
& \CodePolicyWorkflow
& $-7{,}720$ & $-5{,}380$ & $-4{,}110$
& $\mathbf{-5{,}737}$ & $\mathbf{-17.71}$
& $\mathbf{7{,}783}$ & $\mathbf{-5.57}$ \\
& \LLMPolicyWorkflow
& $\mathbf{480}$ & $-11{,}780$ & $-10{,}670$
& $-7{,}323$ & $-22.60$
& $9{,}190$ & $-8.20$ \\
\addlinespace[4pt]

\multirow{2}{*}{DeepSeek-V4.1-Flash}
& \CodePolicyWorkflow
& $-9{,}660$ & $-1{,}800$ & $-9{,}300$
& $\mathbf{-6{,}920}$ & $\mathbf{-21.36}$
& $\mathbf{7{,}193}$ & $-9.66$ \\
& \LLMPolicyWorkflow
& $-1{,}780$ & $-14{,}460$ & $-6{,}030$
& $-7{,}423$ & $-22.91$
& $8{,}280$ & $\mathbf{-8.50}$ \\
\addlinespace[4pt]

\multirow{2}{*}{GLM-5.3}
& \CodePolicyWorkflow
& $-3{,}700$ & $-6{,}310$ & $-6{,}420$
& $-5{,}477$ & $-16.90$
& $\mathbf{5{,}787}$ & $-8.11$ \\
& \LLMPolicyWorkflow
& $-7{,}140$ & $-8{,}280$ & $\mathbf{2{,}230}$
& $\mathbf{-4{,}397}$ & $\mathbf{-13.57}$
& $6{,}583$ & $\mathbf{-3.54}$ \\
\addlinespace[4pt]

\multirow{2}{*}{GLM-5.3-Flash}
& \CodePolicyWorkflow
& $-5{,}150$ & $-5{,}340$ & $-7{,}700$
& $\mathbf{-6{,}063}$ & $\mathbf{-18.71}$
& $\mathbf{7{,}353}$ & $\mathbf{-8.16}$ \\
& \LLMPolicyWorkflow
& $-8{,}700$ & $-11{,}040$ & $-10{,}940$
& $-10{,}227$ & $-31.56$
& $10{,}313$ & $-14.14$ \\
\addlinespace[4pt]

\multirow{2}{*}{GPT-6-Astra}
& \CodePolicyWorkflow
& $-2{,}110$ & $-580$ & $-2{,}150$
& $\mathbf{-1{,}613}$ & $\mathbf{-4.98}$
& $\mathbf{3{,}263}$ & $\mathbf{-1.99}$ \\
& \LLMPolicyWorkflow
& $-4{,}950$ & $-6{,}670$ & $-8{,}510$
& $-6{,}710$ & $-20.71$
& $7{,}577$ & $-7.19$ \\
\addlinespace[4pt]

\multirow{2}{*}{GPT-6-Luna}
& \CodePolicyWorkflow
& $-7{,}840$ & $-9{,}520$ & $-7{,}060$
& $\mathbf{-8{,}140}$ & $\mathbf{-25.12}$
& $\mathbf{8{,}140}$ & $\mathbf{-9.77}$ \\
& \LLMPolicyWorkflow
& $-11{,}640$ & $-8{,}910$ & $-18{,}670$
& $-13{,}073$ & $-40.35$
& $13{,}073$ & $-12.37$ \\
\addlinespace[4pt]

\multirow{2}{*}{GPT-6-Sol}
& \CodePolicyWorkflow
& $-4{,}640$ & $-1{,}880$ & $-1{,}270$
& $\mathbf{-2{,}597}$ & $\mathbf{-8.01}$
& $\mathbf{4{,}503}$ & $\mathbf{-2.95}$ \\
& \LLMPolicyWorkflow
& $-7{,}640$ & $-9{,}770$ & $-8{,}980$
& $-8{,}797$ & $-27.15$
& $9{,}413$ & $-9.95$ \\
\addlinespace[4pt]

\multirow{2}{*}{GPT-6.1-Sol}
& \CodePolicyWorkflow
& $-1{,}080$ & $-5{,}230$ & $-5{,}120$
& $\mathbf{-3{,}810}$ & $\mathbf{-11.76}$
& $\mathbf{4{,}557}$ & $\mathbf{-4.96}$ \\
& \LLMPolicyWorkflow
& $-6{,}430$ & $-7{,}630$ & $-7{,}020$
& $-7{,}027$ & $-21.69$
& $7{,}713$ & $-7.70$ \\
\addlinespace[4pt]

\multirow{2}{*}{MiMo-V2.6-Pro}
& \CodePolicyWorkflow
& $-2{,}880$ & $-3{,}250$ & $\mathbf{12{,}260}$
& $\mathbf{2{,}043}$ & $\mathbf{6.31}$
& $\mathbf{2{,}880}$ & $\mathbf{1.10}$ \\
& \LLMPolicyWorkflow
& $-12{,}520$ & $-5{,}720$ & $-7{,}070$
& $-8{,}437$ & $-26.04$
& $9{,}187$ & $-9.80$ \\

\midrule
\multicolumn{9}{l}{\textbf{Full Kelly, 1\% cap}} \\

\multirow{2}{*}{DeepSeek-V4-Flash}
& \CodePolicyWorkflow
& $-33{,}860$ & $-18{,}410$ & $-15{,}900$
& $\mathbf{-22{,}723}$ & $\mathbf{-15.86}$
& $\mathbf{32{,}683}$ & $\mathbf{-3.19}$ \\
& \LLMPolicyWorkflow
& $\mathbf{1{,}470}$ & $-49{,}790$ & $-55{,}480$
& $-34{,}600$ & $-25.48$
& $63{,}927$ & $-4.09$ \\
\addlinespace[4pt]

\multirow{2}{*}{DeepSeek-V4.1-Flash}
& \CodePolicyWorkflow
& $-59{,}510$ & $-23{,}540$ & $-44{,}990$
& $-42{,}680$ & $-28.74$
& $44{,}613$ & $-8.93$ \\
& \LLMPolicyWorkflow
& $-2{,}950$ & $-55{,}210$ & $-24{,}130$
& $\mathbf{-27{,}430}$ & $\mathbf{-25.94}$
& $\mathbf{41{,}867}$ & $\mathbf{-4.06}$ \\
\addlinespace[4pt]

\multirow{2}{*}{GLM-5.3}
& \CodePolicyWorkflow
& $-33{,}750$ & $-26{,}160$ & $-30{,}040$
& $-29{,}983$ & $-21.09$
& $\mathbf{31{,}583}$ & $-6.22$ \\
& \LLMPolicyWorkflow
& $-46{,}750$ & $-45{,}290$ & $\mathbf{32{,}810}$
& $\mathbf{-19{,}743}$ & $\mathbf{-18.49}$
& $47{,}123$ & $\mathbf{-2.15}$ \\
\addlinespace[4pt]

\multirow{2}{*}{GLM-5.3-Flash}
& \CodePolicyWorkflow
& $-8{,}640$ & $-29{,}280$ & $-41{,}070$
& $\mathbf{-26{,}330}$ & $\mathbf{-19.80}$
& $\mathbf{28{,}770}$ & $\mathbf{-5.20}$ \\
& \LLMPolicyWorkflow
& $-51{,}920$ & $-57{,}910$ & $-53{,}450$
& $-54{,}427$ & $-56.23$
& $54{,}427$ & $-13.34$ \\
\addlinespace[4pt]

\multirow{2}{*}{GPT-6-Astra}
& \CodePolicyWorkflow
& $-3{,}120$ & $\mathbf{15{,}360}$ & $\mathbf{4{,}360}$
& $\mathbf{5{,}533}$ & $\mathbf{4.19}$
& $\mathbf{18{,}637}$ & $\mathbf{0.88}$ \\
& \LLMPolicyWorkflow
& $-37{,}800$ & $-23{,}350$ & $-43{,}020$
& $-34{,}723$ & $-35.26$
& $44{,}813$ & $-5.29$ \\
\addlinespace[4pt]

\multirow{2}{*}{GPT-6-Luna}
& \CodePolicyWorkflow
& $-44{,}420$ & $-31{,}220$ & $-54{,}020$
& $\mathbf{-43{,}220}$ & $-44.57$
& $\mathbf{43{,}220}$ & $-9.16$ \\
& \LLMPolicyWorkflow
& $-56{,}520$ & $-51{,}390$ & $-81{,}080$
& $-62{,}997$ & $\mathbf{-44.20}$
& $62{,}997$ & $\mathbf{-8.26}$ \\
\addlinespace[4pt]

\multirow{2}{*}{GPT-6-Sol}
& \CodePolicyWorkflow
& $-26{,}680$ & $\mathbf{1{,}820}$ & $\mathbf{990}$
& $\mathbf{-7{,}957}$ & $\mathbf{-5.14}$
& $\mathbf{27{,}267}$ & $\mathbf{-1.12}$ \\
& \LLMPolicyWorkflow
& $-46{,}810$ & $-61{,}020$ & $-52{,}310$
& $-53{,}380$ & $-35.44$
& $60{,}723$ & $-8.09$ \\
\addlinespace[4pt]

\multirow{2}{*}{GPT-6.1-Sol}
& \CodePolicyWorkflow
& $-7{,}870$ & $-14{,}340$ & $-19{,}440$
& $\mathbf{-13{,}883}$ & $\mathbf{-10.87}$
& $\mathbf{21{,}993}$ & $\mathbf{-2.44}$ \\
& \LLMPolicyWorkflow
& $-33{,}880$ & $-40{,}820$ & $-36{,}910$
& $-37{,}203$ & $-38.04$
& $47{,}423$ & $-5.70$ \\
\addlinespace[4pt]

\multirow{2}{*}{MiMo-V2.6-Pro}
& \CodePolicyWorkflow
& $-11{,}780$ & $\mathbf{3{,}540}$ & $\mathbf{73{,}080}$
& $\mathbf{21{,}613}$ & $\mathbf{14.25}$
& $\mathbf{19{,}257}$ & $\mathbf{1.46}$ \\
& \LLMPolicyWorkflow
& $-52{,}350$ & $-31{,}340$ & $-27{,}980$
& $-37{,}223$ & $-37.08$
& $49{,}247$ & $-5.80$ \\

\bottomrule
\end{tabular*}
\end{table}

\FloatBarrier

\begin{table}[!t]
\centering
\small
\setlength{\tabcolsep}{4pt}
\renewcommand{\arraystretch}{1.08}
\caption{Mean workflow gross reward, weighted research fees, and net utility per 324-race trajectory (in score-yen; $n=3$). Bold indicates higher gross/net score or lower fees within each model.}
\label{tab:research-net-scores}
\begin{tabular*}{\textwidth}
{@{\extracolsep{\fill}}llrrr@{}}
\toprule
Model & Workflow & Gross & Weighted fees & Net \\
\midrule

\multirow{2}{*}{DeepSeek-V4-Flash}
& \CodePolicyWorkflow
& $13{,}719.29$
& $\mathbf{4{,}596.55}$
& $\mathbf{9{,}122.74}$ \\
& \LLMPolicyWorkflow
& $\mathbf{14{,}395.35}$
& $9{,}355.41$
& $5{,}039.94$ \\
\addlinespace[2pt]

\multirow{2}{*}{DeepSeek-V4.1-Flash}
& \CodePolicyWorkflow
& $11{,}952.78$
& $\mathbf{6{,}180.39}$
& $5{,}772.39$ \\
& \LLMPolicyWorkflow
& $\mathbf{13{,}515.06}$
& $7{,}600.42$
& $\mathbf{5{,}914.64}$ \\
\addlinespace[2pt]

\multirow{2}{*}{GLM-5.3}
& \CodePolicyWorkflow
& $\mathbf{19{,}630.67}$
& $\mathbf{6{,}032.04}$
& $\mathbf{13{,}598.63}$ \\
& \LLMPolicyWorkflow
& $12{,}976.74$
& $10{,}878.60$
& $2{,}098.14$ \\
\addlinespace[2pt]

\multirow{2}{*}{GLM-5.3-Flash}
& \CodePolicyWorkflow
& $\mathbf{18{,}966.73}$
& $\mathbf{7{,}413.87}$
& $\mathbf{11{,}552.86}$ \\
& \LLMPolicyWorkflow
& $12{,}790.81$
& $7{,}742.16$
& $5{,}048.65$ \\
\addlinespace[2pt]

\multirow{2}{*}{GPT-6-Astra}
& \CodePolicyWorkflow
& $\mathbf{24{,}196.43}$
& $\mathbf{4{,}406.25}$
& $\mathbf{19{,}790.17}$ \\
& \LLMPolicyWorkflow
& $22{,}127.04$
& $8{,}593.80$
& $13{,}533.24$ \\
\addlinespace[2pt]

\multirow{2}{*}{GPT-6-Luna}
& \CodePolicyWorkflow
& $\mathbf{11{,}912.35}$
& $\mathbf{2{,}595.57}$
& $\mathbf{9{,}316.78}$ \\
& \LLMPolicyWorkflow
& $5{,}419.79$
& $4{,}008.41$
& $1{,}411.38$ \\
\addlinespace[2pt]

\multirow{2}{*}{GPT-6-Sol}
& \CodePolicyWorkflow
& $\mathbf{22{,}071.07}$
& $\mathbf{5{,}457.72}$
& $\mathbf{16{,}613.35}$ \\
& \LLMPolicyWorkflow
& $16{,}503.75$
& $6{,}537.17$
& $9{,}966.58$ \\
\addlinespace[2pt]

\multirow{2}{*}{GPT-6.1-Sol}
& \CodePolicyWorkflow
& $\mathbf{23{,}748.71}$
& $\mathbf{4{,}535.80}$
& $\mathbf{19{,}212.91}$ \\
& \LLMPolicyWorkflow
& $22{,}608.45$
& $10{,}371.01$
& $12{,}237.44$ \\
\addlinespace[2pt]

\multirow{2}{*}{MiMo-V2.6-Pro}
& \CodePolicyWorkflow
& $\mathbf{20{,}509.72}$
& $\mathbf{5{,}225.68}$
& $\mathbf{15{,}284.05}$ \\
& \LLMPolicyWorkflow
& $14{,}161.38$
& $9{,}570.57$
& $4{,}590.81$ \\

\bottomrule
\end{tabular*}
\end{table}

\subsection{Research costs and synthetic utility}
\paragraph{Code-as-policy fee decomposition.}
Among the top six code-as-policy configurations in Table~\ref{tab:main}, research fees range from 12.2 to 15.5 score-yen per race, dominated by warmup exploration and per-race execution data reads. Incorporating fees and hit bonuses can reorder closely matched predictors: for example, Claude Opus 5.5 has a slightly lower log gain than GPT-6-Sol ($0.453$ vs.\ $0.471$ nats) but achieves a higher net utility ($+53.1$ vs.\ $+50.2$ score-yen/race) due to lower per-race research fees ($12.2$ vs.\ $16.7$ score-yen).

\paragraph{Workflow score decomposition and Pareto dominance.}
Table~\ref{tab:research-net-scores} and Figure~\ref{fig:workflow-research-scores} compare mean gross reward ($\mathrm{Gross}_{\mathrm{cmp}}$), weighted research fees ($C$), and net score ($\mathrm{Net}_{\mathrm{cmp}}$) per 324-race trajectory across the nine shared models. \emph{code-as-policy} incurs lower research fees on all nine models and achieves higher net utility on eight. On seven models (all except the two DeepSeek variants), code-as-policy achieves both higher gross reward and lower research fees---strictly dominating LLM-as-policy under any positive tariff multiplier. LLM-as-policy achieves higher gross reward on DeepSeek-V4-Flash and DeepSeek-V4.1-Flash, but its extra research fees outweigh the gross gain on DeepSeek-V4-Flash, leaving only DeepSeek-V4.1-Flash with a slight net-score edge ($+142.26$ score-yen over 324 races).

\clearpage
\section{Prompt Templates and Submission Interfaces}
\label{app:prompt}
In the code-as-policy workflow, the LLM is invoked during initial exploration and daily reflection (with at most one repair session per phase), whereas for each race the framework executes the saved Python predictor directly without an LLM call. In the LLM-as-policy workflow, the LLM is invoked to build the initial evidence plan, to update it after each race day, and once per race to forecast (with at most one repair request), whereas the evidence plan itself runs without an LLM call. Templates for both workflows and their provenance are given below.

\definecolor{RBPromptHeader}{HTML}{666666}
\definecolor{RBPromptBackground}{HTML}{F2F2F2}
\newcommand{\rbprompttype}{\fontfamily{lmtt}\fontsize{9}{11}\selectfont}
\newtcolorbox{rbprompt}[2][]{
  enhanced,
  colframe=RBPromptHeader, colback=RBPromptBackground,
  colbacktitle=RBPromptHeader, coltitle=white,
  boxrule=1.2pt, arc=4pt, outer arc=4pt,
  boxsep=0pt, left=16pt, right=16pt, top=9pt, bottom=9pt,
  lefttitle=16pt, righttitle=16pt, toptitle=3pt, bottomtitle=3pt,
  fonttitle=\rmfamily\bfseries\fontsize{12}{14}\selectfont,
  fontupper=\rbprompttype,
  before upper={\raggedright\setlength{\parindent}{0pt}\setlength{\parskip}{6pt}},
  before skip=8pt, after skip=8pt,
  title={#2},
  #1
}

\subsection{\texorpdfstring{\textsc{Code-as-policy phase prompts}}{Code-as-policy phase prompts}}
\label{app:prompt-examples}
The reference system prompt was regenerated from the archived implementation and checked against the stored specification. This check covered the reference template and excluded historical client wrappers and run-specific messages.

The prompts below use the paper's code-as-policy terminology.
\begin{rbprompt}[breakable,title after break={Code-as-policy system prompt (continued)}]{Code-as-policy system prompt}
\begin{lstlisting}[language={},basicstyle=\rbprompttype,
breaklines=true,breakatwhitespace=false,breakautoindent=false,
columns=fullflexible,keepspaces=true,showstringspaces=false,
frame=none,backgroundcolor=\color{RBPromptBackground},
aboveskip=0pt,belowskip=0pt,literate={'}{{\textquoteright}}1]
Task
1. Create and improve strategy.py defining predict(sdk, context). The framework executes and seals every race prediction. Model sessions occur during initial exploration, daily reflection and at most one repair per phase. No odds, popularity, payouts for current or past races, JRDB indices, marks, predictions, aptitude or class ratings, track-bias figures or free text are available. Unknown values are null with missing reasons.
2. Feedback schema: items[] contains this day's races only; race_id is its identifier. top1_hits/top1_n are this day's champion hits/scored races; top3_hits/top3_n are this day's correct top-three choices/eligible selected choices. These are daily counts, not cumulative. Each item has status (ok, cancelled, no_valid_runners or missing_results) and revenue_yen (that race's gross score or null), revenue_log_yen (log component or null when unscored), and revenue_hit_yen (hit component, zero when ineligible or unscored; always zero for score-yen-v1). hits contains the daily counts; ledger is cumulative. Ledger fields: currency, tariff and multiplier identify units; cost_yen is incurred cost; settled_revenue_yen is booked gross; net_score is booked gross minus cost; remaining_days counts unpredicted days; monetary_cap=null. Ledger hit_bonus_yen is the configured bonus; revenue_log_yen/revenue_hit_yen are the cumulative booked components.

Time contract
1. For every race of day D, predict runs at the previous evening cutoff, 21:00 JST. Visible inputs are that card, workouts and permitted labels released up to that cutoff. Day D results release at 20:30 JST; later objective review fields release at their review publication time. Reflection runs after the day closes and before day D+1 cards exist (21:00 JST). Attached form is frozen at its target race cutoff even when queried later. context['seed'] is constant per trajectory and is the only sanctioned randomness source; seed any random generator from it. Initial support labels require review release; online finish/abnormal labels become accessible after day close.

Code-as-policy interface and validation
1. Define callable predict(sdk, context) in strategy.py. Return a JSON object containing only ranking and probabilities; ranking is a list of every entrant number string exactly once, with no omissions, extras or duplicates. Probabilities are mandatory. When supplied, probabilities map every entrant to a finite number in [0,1]; their sum must be within absolute tolerance 0.02 of 1 (positive mass required). Accepted rounding is normalized; sums within 1e-12 are retained. predict must not write files. sdk.call(tool_name, **arguments) returns the same JSON as the direct tool. sdk.history(...) is a paging generator. There are no per-tool methods such as sdk.card. SDK data calls inside predict are the only way a strategy sees the race data; context supplies only identifiers, runner numbers and seed. 
Validation checklist: Python syntax and two-argument predict declaration on save; callable predict at execution; JSON output schema; full ranking; probability domain and normalization; no workspace changes; per-race calls, evidence and CPU caps; job wall time, RSS, storage, artifact, open-file and output/IPC ceilings where configured. Imports and uncaught exceptions can fail a job.
Failure matrix: save syntax/artifact rejection -> tool error, correct and save again; candidate execution/interface/resource failure -> candidate rejected, previous valid version retained, one repair session; prediction execution/interface/resource failure -> try the previous valid version once, then uniform. Storage quota or an unresolved interruption stops the run for inspection. Missing strategy or mode=uniform -> deterministic fallback ranking with uniform probabilities.
Each of up to three initial-support validation races executes twice. Differing ranking or probabilities sets nondeterministic_output=true in last_validation and prompts the same one repair session; it does not invalidate the code-as-policy predictor. Validation uses no predictive-accuracy threshold.
Uninformative-output rule: all returned probabilities are exactly equal. If this holds on every validation race, silent_uniform requests the one repair session, regardless of evidence read. The code-as-policy predictor stays valid if kept, including after repeated uninformative output.

Scoring
1. Maximize net_score = sum_races [S*ln(n_r*pbar_r) + I(delivery=predictor)*hit_bonus_yen*(h_r-1/n_r)] - sum(evidence fees + call fees + CPU fees) over the entire trajectory. After removing scratched runners from the sealed ranking, renormalize their remaining probability mass; if it is zero use uniform. n_r is the number that actually started. Smooth each remaining probability: p_smoothed=(1-lambda)*p+lambda/n_r. pbar_r is the mean smoothed probability over ALL official winners (dead heat: every official winner counts). Tariff score-yen-v2: hit_bonus_yen=100. Ranking now has a scoring role: h_r=1 if the top-ranked runner that actually started is any official winner, else 0. The hit bonus requires a validated code-as-policy predictor and sealed delivery=predictor; I is 1 only for that delivery, otherwise 0. explicit_uniform, no_strategy, silent_uniform and execution_fallback receive exactly zero hit bonus, including fallback to an older valid version. At n=14 and hit_bonus_yen=100, a hit pays +92.9 and a miss pays -7.1, in addition to the log part. The hit bonus is independent of submitted probabilities; with one official winner, a uniformly random top choice has expected bonus zero. revenue_log_yen and revenue_hit_yen are booked separately and sum to revenue_yen. Invalid/cancelled results book zero gross score. Each day's revenue is booked when that day closes; costs are booked when incurred. The final value is the ledger after the last day and its reflection close. S=100; lambda=0.02. At S=100, lambda=0.02, n=14 and winner probability 0.10, p_smoothed=0.0994285714 and log revenue=33.1; uniform gives 0. For beliefs q with a feasible interior optimum, the optimal report is p=(q-lambda/n)/(1-lambda), within O(lambda) of belief; otherwise the constrained optimum has boundary components. Fees do not depend on the reported probabilities.

Costs and units
1. One evidence cell is one scalar fact slot delivered, null included; identifiers/day indices count, duplicate metadata and missing-reason strings do not. Statistics/coverage meter underlying scalar evidence inspected. field_yen=0.01 score-yen/cell; call_yen=0.5 per data-tool call; cpu_yen=0.2 per CPU-second (user+system) of the isolated process, including failed code. All three rates are multiplied by 0.5. Data tools: card, connections, coverage, form, history, horse_history, results, statistics, training. Control tools are free of fees; compute adds CPU and nested data-tool fees. Tool steps and response bytes still count for controls. From the live field catalogue, a card with all three entity identifiers per runner delivers 25 + 43*n cells: n=14 gives 627 cells, approximately 3.39 score-yen including one call (excluding CPU); omitted entity identifiers reduce the count. Score-yen is a scoring unit, not real currency. No monetary cap or requirement to spend; uniform without any incurred costs has net_score zero. Within one agent's own trajectory (its initial exploration is shared by that agent's own arms only), the same tool and canonical arguments (excluding max_spend) with an identical response digest is a repeat: evidence fee and charged_cells are zero and no session or per-race evidence allowance is consumed. Call fees and call/byte counters are unchanged. A changed response digest is charged in full. Only digests are retained, never data pages.

Allowances
1. Phase | Allowances | Shared or separate | At cap
initial exploration | warmup_calls=1000, warmup_evidence_cells=400000, warmup_bytes=67108864, warmup_model_calls=40; main=35, repair reserve=5 | Direct tools and nested SDK calls share this phase; repair shares remaining data allowances. | Data calls reject recoverably; model-call or turn cap ends this session.
daily reflection | batch_calls=1000, batch_evidence_cells=100000, batch_bytes=33554432, batch_model_calls=20; main=15, repair reserve=5 | Direct tools and nested SDK calls share this phase; repair shares remaining data allowances. | Data calls reject recoverably; model-call or turn cap ends this session.
repair | repair_model_calls=5 | One repair session per phase; reserved calls plus all unused main calls, within the original phase cap. | Keep the last valid code-as-policy predictor or uniform; no second repair session.
per-race prediction and validation job | strategy_calls=40, strategy_evidence_cells=5000, strategy_cpu_seconds=120 | Separate from sessions; each execution has fresh caps (including each validation repeat). Outer compute and all SDK calls count. | Job fails; candidate rejection or prediction fallback applies.
2. Direct responses exceeding response_bytes=1048576 (including a reserved 2048-byte envelope) are rejected uncharged, with no data or call fee. A smaller remaining model context can lower this ceiling. max_page_size=8; default_page_size=2.
3. batch_seconds limits cumulative session compute wall-time within a phase, including repair; prediction and validation wall-time is separate. null means no ceiling.
4. Additional session limits: {"batch_bytes": 33554432, "batch_calls": 1000, "batch_evidence_cells": 100000, "batch_model_calls": 20, "batch_seconds": null, "context_bytes": 262144, "default_page_size": 2, "feedback_memory_bytes": 16384, "feedback_memory_items": 32, "format_repairs": 1, "max_page_size": 8, "model_output_tokens": null, "model_turns": 64, "repair_model_calls": 5, "response_bytes": 1048576, "strategy_calls": 40, "strategy_cpu_seconds": 120, "strategy_evidence_cells": 5000, "warmup_bytes": 67108864, "warmup_calls": 1000, "warmup_evidence_cells": 400000, "warmup_model_calls": 40}

Sessions and persistence
1. Every compute starts fresh with injected sdk. Assign result to a small JSON value. Variables do not persist. Persist only small code, concise NOTES.md and fitted parameters; historical samples, race packages and training arrays on disk are rejected and rolled back. predict cannot modify files. Files written during a session persist only if save_strategy is called at least once (saving NOTES.md suffices); otherwise restore the session-start workspace. Current source, notes and last_validation are shown in the next session. Watch _budget; save when model calls <=3 or a data allowance <=20%. Oldest complete tool rounds are dropped first; the task message, strategy source and notes are never dropped (oversize initial context ends the session; source inclusion has an explicit truncation manifest).

Execution environment
1. {"hash_seed": 0, "imports": {"numpy": true, "pandas": false, "scipy": true, "sklearn": true, "statsmodels": false}, "isolation": "no network, no subprocess, workspace-only file access", "numerical_threads": 4, "python_version": "3.13.2", "runtime_exception": "Read-only interpreter and installed libraries are available.", "system": "Darwin", "system_release": "27.0.0"}
Machine limits: {"artifact_bytes": null, "job_seconds": 1800, "open_files": null, "rss_bytes": 10737418240, "storage_bytes": 5368709120}. rss_bytes is process resident memory; job_seconds is wall time per execution; storage_bytes is total managed trajectory disk usage including checkpoints; artifact_bytes bounds a file; open_files bounds open descriptors. null means no framework ceiling. CPU is still billed for every second used.

Glossary
1. Initial support: races before window day 0 whose labels were released by the initial clock. Trajectory: one agent's exploration and scheduled prediction days. Entrant: declared runner identified by number. Cutoff: latest permitted input time. Sealed: immutable saved prediction. Validation: interface/resource checks on support cases, without model selection by outcomes. Reflection: optional code-as-policy update after a day closes. net_score: cumulative gross score minus incurred fees.
\end{lstlisting}
\end{rbprompt}

The frozen compatible-API client uses these phase instructions:
\begin{rbprompt}{Initial construction.}
``Initial exploration. Build an initial strategy from scratch and save it. No predictions yet.''
\end{rbprompt}
\begin{rbprompt}{Daily revision.}
``Day complete. You may change strategy or leave it unchanged. No new predictions until next batch.''
\end{rbprompt}
\begin{rbprompt}{Repair.}
``Repair the candidate code-as-policy predictor with remaining session resources, or choose uniform.''
\end{rbprompt}

\subsection{\texorpdfstring{\textsc{LLM-as-policy prompts}}{LLM-as-policy prompts}}
\label{app:prompt-llm-policy}
The fixed strings below were extracted statically from the archived source of the DeepSeek-V4-Flash R1 revisable and \Frozen{} deployment. The revisable and \Frozen{} arms share the forecasting and repair prompts; \Frozen{} reuses the exact $P_0$ and runs no updater. The forecasting prompt states the gated objective used during the run; the workflow comparison rescores saved forecasts without the gate (Appendix~\ref{app:acquisition-scoreyen}).

\begin{rbprompt}[breakable,title after break={Per-race forecasting system prompt (continued)}]{Per-race forecasting system prompt}
\begin{lstlisting}[language={},basicstyle=\rbprompttype,breaklines=true,breakatwhitespace=false,breakautoindent=false,columns=fullflexible,keepspaces=true,showstringspaces=false,frame=none,backgroundcolor=\color{RBPromptBackground},aboveskip=0pt,belowskip=0pt]
You predict exactly one horse race from its legal basic race card and EvidencePackage.
Use exactly one offered native submit function. Never return prose or a batch of calls.
All evidence is as-of the frozen pre-race cutoff. Never request future results or another race.
Recent history and condition stats use the native latest-five-start history. They are not lifetime statistics.
Null/missing means unknown, not zero. Opaque categorical codes have no assumed codebook.
Submit a complete ranking and finite probabilities for every entrant summing to 1.
The aligned score is 100*ln(n*(0.98*p_winner+0.02/n)) plus
100*(Top1_hit-1/n) for a valid non-uniform prediction, minus 0.5 times raw research cost.
Agent uniform and framework fallback have zero Top1 bonus. Service fees already incurred remain payable.
Research is complete. Only submit is available. Return the native function call only.
\end{lstlisting}
\end{rbprompt}
The user message is a JSON object with fields \texttt{mode} (\texttt{"prediction"}), \texttt{basic\_race\_card}, \texttt{EvidencePackage} (compact semantic JSON), and \texttt{ledger}. Only the \texttt{submit} tool is offered, with \texttt{tool\_choice} forced to it.

\begin{rbprompt}{Repair suffix (appended to the forecasting system prompt)}
\begin{lstlisting}[language={},basicstyle=\rbprompttype,breaklines=true,columns=fullflexible,keepspaces=true,showstringspaces=false,frame=none,backgroundcolor=\color{RBPromptBackground},aboveskip=0pt,belowskip=0pt]
Research is now closed.

Using only the race information and evidence already available in this request, submit the final prediction for this race.

No further research tools are available.
Return exactly one valid native submit action.
\end{lstlisting}
\end{rbprompt}

\begin{rbprompt}[breakable,title after break={Initial evidence-plan construction (continued)}]{Initial evidence-plan construction (warmup) system prompt, task part}
\begin{lstlisting}[language={},basicstyle=\rbprompttype,breaklines=true,breakatwhitespace=false,breakautoindent=false,columns=fullflexible,keepspaces=true,showstringspaces=false,frame=none,backgroundcolor=\color{RBPromptBackground},aboveskip=0pt,belowskip=0pt]
Build a reusable information-acquisition program using legal historical support. Improve future information acquisition efficiency, not horse outcome prediction. The program will be reused for future races. Inspect the contract and schema; explore historical evidence and released outcomes; use compute to analyze it; validate and save an acquisition-only program, then finish. Historical labels are for analysis only: the saved program has no results, prediction or submit capability. No predictive parameters, auxiliary files or analysis workspace are transferred to the runtime. Do not guess unavailable facts. Use sdk.call(name, **arguments) inside compute; assign its final output to result. Each compute starts fresh; only its own flat workspace persists. No network or external file access. All data and compute are metered. No prebuilt acquisition policy is supplied.
\end{lstlisting}
\end{rbprompt}
The warmup system prompt continues with the acquisition contract below. Its first user message supplies the task, support clock, relative day origin, objective ($B=100$, cost multiplier 0.5), tariff, limits, and an allowance of 35 normal requests plus 5 repair requests.

\begin{rbprompt}[breakable,title after break={Acquisition contract (continued)}]{Acquisition contract (shared by warmup and updater)}
\begin{lstlisting}[language={},basicstyle=\rbprompttype,breaklines=true,breakatwhitespace=false,breakautoindent=false,columns=fullflexible,keepspaces=true,showstringspaces=false,frame=none,backgroundcolor=\color{RBPromptBackground},aboveskip=0pt,belowskip=0pt]
Define collect_evidence(sdk, context), with optional deterministic helper functions.
context contains only race_id, day_index, feature_asof, field_size and runners[{number}].
Use sdk.state() for a copied InformationState; sdk.state(horse_number) for that runner.
InformationState has query_count, raw_research_cost, weighted_research_cost,
domain_coverage, domains and runners. Runner state has number, stopped, domains.
Domain fields: queried, query_count, queried_depth, records_returned, missing_count, observed_slots, missing_ratio, feature_asof, freshness_days, available_history_start_count, deeper_available, same_surface_records, same_distance_records, sample_starts, labelled_starts, matching_groups.
Unknown unpurchased availability/freshness/missingness is None. Do not infer it as zero.
sdk.call("recent_history", horse_number=number, depth=1|3|5); default depth=5.
sdk.call("training", horse_number=number); sdk.call("condition_stats", horse_number=number).
sdk.call("track_context") exactly once, explicitly. No depth or repeat for track_context.
sdk.call returns quality-only runner/race InformationState; raw observations stay in FullEvidencePackage.
History expansions purchase only newly acquired starts/cells under the native tariff.
Each actual native operation retains its native call fee; cached repeats are local projections.
Training is once-per-race whole native acquisition followed by free local horse projections.
condition_stats uses the shared owned history cache, acquiring only missing starts up to five.
The initial card and first training acquisition retain their distinct native tariff identities.
Frame distinguishes logical policy actions, native acquisition operations, newly acquired cells and local projections.
Use sdk.stop_horse(number) or sdk.stop_domain(domain, number) for explicit stopping;
break/continue and omission of additional calls also stop acquisition. No calls after explicit stopping.
Finish with return sdk.package(), an opaque token. No policy-authored predictor payload.
Supported: bounded for loops over runners/list/range, if, comparisons, arithmetic,
local bindings, len/range/sorted/min/max/abs, dict.get, nonrecursive helpers.
No imports, globals, mutation, while, recursion, arbitrary attributes, I/O, model
calls, results, prediction, rank, finish, wins/top3, rates, strength, score or probability.
Rules/thresholds/ordering/stopping are generated by the updater, not the framework.
The research program should decide whether additional information is necessary,
not which horse is likely to win. Optimize future information acquisition efficiency.
Historical acquisition/coverage/cost/stopping and legally released race-level
metrics may inform updater analysis; no horse prediction/ranking/scoring rule may be saved.
Observed feedback relationships are descriptive association, not causal value.
\end{lstlisting}
\end{rbprompt}

\begin{rbprompt}{Daily updater system prompt (contract inserted where marked)}
\begin{lstlisting}[language={},basicstyle=\rbprompttype,breaklines=true,columns=fullflexible,keepspaces=true,showstringspaces=false,frame=none,backgroundcolor=\color{RBPromptBackground},aboveskip=0pt,belowskip=0pt]
Construct deterministic information-acquisition code, not predictions. [acquisition contract]

Use the provided tools in a multi-turn inspect/validate/save/finish session. Synthetic fixtures contain no real outcomes. Invalid updates retain the last valid program. Observed feedback associations are not causal contributions.
\end{lstlisting}
\end{rbprompt}
Each updater turn permits one tool call. Appendix~\ref{app:active-research} describes the update context and validation rules.

\subsection{\texorpdfstring{\textsc{Task context and workflow differences}}{Task context and workflow differences}}
The code-as-policy task object supplies the horizon, remaining days, current predictor and notes, validation feedback, recent errors, and the cost ledger. Revision messages add released daily feedback. Repair messages add a validation notice, including invalid, uniform, or nondeterministic candidate output. Fields are populated at the replay clock. \Frozen{} skips \Live{}'s daily revision sessions while retaining time-varying legal inputs.

\subsection{\texorpdfstring{\textsc{Minimal submission example}}{Minimal submission example}}
The following uniform predictor comes from the framework's SDK help, with line wrapping adjusted. It shows data access and the required return schema. R0 uses a separate seeded ranking implementation.
\begin{rbprompt}{Minimal submission example}
\begin{lstlisting}[language=Python,basicstyle=\rbprompttype,
breaklines=true,columns=fullflexible,keepspaces=true,showstringspaces=false,
frame=none,backgroundcolor=\color{RBPromptBackground},aboveskip=0pt,belowskip=0pt]
def predict(sdk, context):
    card = sdk.call(
        'card', race_id=context['race_id'])
    numbers = [runner['number']
               for runner in card['runners']]
    return {
        'ranking': numbers,
        'probabilities': {
            n: 1 / len(numbers)
            for n in numbers}}
\end{lstlisting}
\end{rbprompt}
This output is classified as \texttt{silent\_uniform} and receives no predictor-delivery hit bonus.

\begingroup
\raggedbottom
\section{\texorpdfstring{\textsc{Case Studies of Executed Strategies}}{Case Studies of Executed Strategies}}
\label{app:case-studies}
The code-as-policy cases each cover 324 races and compare daily revision (\Live{}) with the matched frozen initial predictor (\Frozen{}). The LLM-as-policy case compares frozen and revisable evidence plans in round 1 of DeepSeek-V4-Flash. Condensed pseudocode summarizes the executed predictors, saved parameters, and query schedules. Here, $n$ is the field size. Code-as-policy probability transformations precede evaluator smoothing. Reflection excerpts reproduce model replies verbatim. Ellipses mark omissions.

\definecolor{RBCaseHeader}{HTML}{666666}
\definecolor{RBCaseBackground}{HTML}{F2F2F2}
\newcommand{\rbcasetype}{\fontfamily{lmtt}\fontsize{9}{11}\selectfont}
\newtcolorbox{rbcase}[1]{
  enhanced,
  colframe=RBCaseHeader, colback=RBCaseBackground,
  colbacktitle=RBCaseHeader, coltitle=white,
  boxrule=1.2pt, arc=4pt, outer arc=4pt,
  boxsep=0pt, left=16pt, right=16pt, top=9pt, bottom=9pt,
  lefttitle=16pt, righttitle=16pt, toptitle=3pt, bottomtitle=3pt,
  fonttitle=\rmfamily\bfseries\fontsize{12}{14}\selectfont,
  fontupper=\rbcasetype,
  before upper={\raggedright\setlength{\parindent}{0pt}\setlength{\parskip}{6pt}},
  before skip=8pt, after skip=8pt,
  title={#1}
}
\newcommand{\rbphase}[1]{\par\textbf{#1.}\ }
\lstdefinestyle{rbcasecode}{
  basicstyle=\rbcasetype, columns=fullflexible, keepspaces=true,
  showstringspaces=false, breaklines=true, frame=none,
  backgroundcolor=\color{RBCaseBackground},
  xleftmargin=0pt, xrightmargin=0pt, aboveskip=6pt, belowskip=6pt
}

\subsection{\texorpdfstring{\textsc{Initial strategy construction}}{Initial strategy construction}}
\phantomsection\label{app:case-astra}
\begin{rbcase}{GPT-6-Astra: a fixed linear predictor}
\textit{Selected trajectory: repeat 3. One executed predictor version.}

\rbphase{Initial strategy}
The predictor constructs 42 features from the race card, the two most recent starts, and jockey/trainer statistics. Recent-form summaries weight the latest start by 0.65 and the preceding start by 0.35. A linear score uses fitted feature means, scales, and coefficients. It converts scores into probabilities and mixes in a small uniform component.
\begin{lstlisting}[style=rbcasecode]
x = features(card, form(k=2), connections)  # 42 features
z = (x - fitted_mean) / fitted_scale
score = z @ fitted_weights
p = 0.98 * softmax(score) + 0.02 / n
ranking = descending_order(p)
\end{lstlisting}

\rbphase{Reflection after Day~1 (excerpt)}
``This supports continuing without costly retraining or overreacting to one day.''

\rbphase{Corresponding execution}
The initial artifact remains the executed predictor on days 1--10. It still reads the current permitted inputs for each race, using three data calls. All 324 probability vectors and complete rankings match \Frozen{}.

\textbf{Results (\Frozen{} $\rightarrow$ \Live{}).} Log gain: $0.517 \rightarrow 0.517$ nats. Top-1: $29.6\% \rightarrow 29.6\%$.

\end{rbcase}

\begin{rbcase}{\phantomsection\label{app:case-sol}GPT-6.1-Sol: a linear and tree score ensemble}
\textit{Selected trajectory: repeat 1. One executed predictor version.}

\rbphase{Initial strategy}
The predictor constructs 86 features from the race card, two recent starts, and connection statistics. It centers features within each race, divides by fitted scales, and clips the transformed values to $[-6,6]$. The saved predictor combines a linear score with an ensemble of 80 depth-two decision trees. Inspection of the fitted trees confirms that all 80 are nonconstant.

\rbphase{Executed prediction rule}
The tree contribution sums the trees with a learning rate of 0.04. Linear and tree scores receive weights 0.7 and 0.3, respectively. Their combined score is multiplied by 1.2 before softmax, followed by a 3\% uniform mixture.
\begin{lstlisting}[style=rbcasecode]
x = features(card, form(k=2), connections)  # 86 features
z = clip((x - race_mean(x)) / fitted_scale, -6, 6)
linear = z @ fitted_weights
trees = 0.04 * sum(tree(z) for tree in saved_trees)  # 80 trees
score = 1.2 * (0.7 * linear + 0.3 * trees)
p = 0.97 * softmax(score) + 0.03 / n
\end{lstlisting}
The combination occurs at the score level, before probability normalization.

\rbphase{Execution and revision trace}
The same fitted predictor is used throughout days 1--10. Its feature inputs are updated for each race, but the executed parameters and architecture remain fixed. \Live{} and \Frozen{} produce identical forecasts throughout the window.

\textbf{Results (\Frozen{} $\rightarrow$ \Live{}).} Log gain: $0.527 \rightarrow 0.527$ nats. Top-1: $29.6\% \rightarrow 29.6\%$.

\end{rbcase}

\subsection{\texorpdfstring{\textsc{Confidence adjustment}}{Confidence adjustment}}
\phantomsection\label{app:case-kimi}
\begin{rbcase}{Kimi-K3: temperature updates with fixed rankings}
\textit{Selected trajectory: repeat 5. Seven executed source snapshots.}

\rbphase{Initial strategy and observed feedback}
The initial predictor uses 16 fixed feature weights and temperature $T=1$. Its first day has mean log gain $-0.243$ nats. Subsequent saved predictors change the temperature while retaining the feature weights and ranking logic.

\rbphase{Reflection after Day~1 (excerpt)}
``Change made: kept the ranking model unchanged, but flattened the probability calibration by raising \texttt{TEMP} from \texttt{1.0} to \texttt{3.0}.''

\rbphase{Corresponding execution}
The temperature is 3 on days 2--5 and 2 on days 6--10. The core confidence transformation is shown below. Raising the positive temperature flattens softmax probabilities while preserving the ordering of the underlying scores.
\begin{lstlisting}[style=rbcasecode]
score = features @ fixed_weights  # same 16 weights
temperature_by_day = [1, 3, 3, 3, 3, 2, 2, 2, 2, 2]
T = temperature_by_day[day - 1]
q = softmax(score / T)  # core temperature transformation
\end{lstlisting}
The seven source snapshots contain only three distinct parsed code structures. The final four source changes add comments without changing executable code.

Compared with \Frozen{}, \Live{} changes 300 probability vectors but no complete ranking or top pick. Mean top-pick probability falls from 53.9\% in \Frozen{} to 29.3\% in \Live{} on the same 324 races.
\textbf{Results (\Frozen{} $\rightarrow$ \Live{}).} Log gain: $0.093 \rightarrow 0.383$ nats. Top-1: $25.3\% \rightarrow 25.3\%$.

\end{rbcase}

\subsection{\texorpdfstring{\textsc{Feature and weight revision}}{Feature and weight revision}}
\phantomsection\label{app:case-deepseek}
\begin{rbcase}{DeepSeek-V4.1-Flash: feature and probability revisions}
\textit{Selected trajectory: repeat 3. Seven executed source snapshots.}

\rbphase{Strategy construction}
The saved predictors use weighted feature scores. Their feature counts change from 12 on day 1 to 7 on days 2--3, 10 on days 4--6, 14 on day 7, and 25 on days 8--10. The final feature set combines 14 card, connection, and workout features with 11 recent-form features, querying up to five previous starts.

\rbphase{Executed final strategy}
Missing numerical values are filled with the observed mean within the race. Features are then standardized within that race and combined linearly. The predictor ranks runners by score. Its submitted probabilities apply a clipped adjustment after softmax.
\begin{lstlisting}[style=rbcasecode]
x = features(card, connections, form(k=5))  # 25 features
x = fill_missing_with_race_mean(x)
z = standardize_within_race(x)
score = z @ weights
q = softmax(score)
p = normalize(maximum(q - 0.02 / n, 0))
ranking = descending_order(score)
\end{lstlisting}
The final transformation offsets evaluator smoothing only when clipping is inactive. With clipping, it is not an exact inverse.

\rbphase{Revision trace and paired outcome}
Seven source snapshots correspond to six distinct parsed code structures. The last two snapshots have the same parsed structure.
\textbf{Results (\Frozen{} $\rightarrow$ \Live{}).} Log gain: $0.288 \rightarrow 0.230$ nats. Top-1: $22.5\% \rightarrow 19.8\%$.

\end{rbcase}

\phantomsection\label{app:case-fable}
\begin{rbcase}{Claude Fable 5.1: early weight updates and later calibration}
\textit{Selected trajectory: repeat 3. Nine executed source snapshots.}

\rbphase{Initial strategy}
The initial predictor uses 28 features and a saved coefficient vector. Both arms use this common predictor for the first 24 races. The prediction functions retain the same structure through the early revisions, while the saved weights change.

\rbphase{Reflection after Day~3 (excerpt)}
``\ldots{} run a CV'd refit, and update only if it holds up.''

\rbphase{Corresponding execution}
Before each of days 2, 3, and 4, 27 of the 28 weight entries change relative to the preceding version. From day 5 onward, the weight vector stays fixed. Later saved versions change calibration or metadata. The parameter history is:
\begin{lstlisting}[style=rbcasecode]
day 1:      w0
day 2:      w1  # 27 of 28 entries differ from w0
day 3:      w2  # 27 of 28 entries differ from w1
day 4:      w3  # 27 of 28 entries differ from w2
days 5-10:  w3  # later calibration / metadata changes
\end{lstlisting}

\Live{} changes 70 top picks and 298 complete rankings relative to \Frozen{}. Its probability vectors differ on 300 races. All 324 races have predictor delivery. The trajectory records ten provider failures.
\textbf{Results (\Frozen{} $\rightarrow$ \Live{}).} Log gain: $0.471 \rightarrow 0.486$ nats. Top-1: $27.8\% \rightarrow 30.2\%$.

\end{rbcase}

\subsection{\texorpdfstring{\textsc{Information acquisition in both workflows}}{Information acquisition in both workflows}}
\phantomsection\label{app:case-terra}
\begin{rbcase}{GPT-5.6-Terra (code-as-policy): switching to a card-only predictor}
\textit{Selected trajectory: repeat 1. Two executed predictor versions.}

\rbphase{Initial strategy}
The first predictor reads the race card and recent form for every prediction, using two data calls. This is also the initial predictor retained by \Frozen{}.

\rbphase{Reflection after Day~3 (excerpt)}
``Change: removed the costly separate \texttt{form(k=3)} call while retaining card, pedigree, and embedded workout features. This reduces per-race evidence cost substantially.''

\rbphase{Corresponding execution}
From day 4, \Live{} uses a shorter card-only predictor and stops requesting recent form. Data calls fall to one per prediction. The source shrinks from 123 to 70 lines, and the revised version handles the remaining 252 races. The revision also changes the prediction code.
\begin{lstlisting}[style=rbcasecode]
days 1-3:
    card, form = read_card(), read_recent_form()
    prediction = initial_predictor(card, form)
days 4-10:
    card = read_card()
    prediction = revised_predictor(card)
\end{lstlisting}

\rbphase{Paired outcome and accounting}
\Live{} changes 102 top picks relative to \Frozen{}. Across the full trajectory, protocol fees fall by 3.53 score-yen per race, but gross forecast reward falls by 17.97. Net utility therefore falls by 14.44 score-yen per race.
\textbf{Results (\Frozen{} $\rightarrow$ \Live{}).} Log gain: $0.333 \rightarrow 0.187$ nats. Top-1: $20.1\% \rightarrow 16.7\%$. Net score: $30.51 \rightarrow 16.07$ score-yen/race.

\end{rbcase}

\phantomsection\label{app:case-acquisition}
\begin{rbcase}{DeepSeek-V4-Flash, round 1 (LLM-as-policy): changing the query schedule}
\textit{Selected comparison: frozen and revisable evidence plans, 324 races per arm.}

\rbphase{Initial strategy}
Both arms start with the same evidence plan $P_0$. It collects evidence for a fresh per-race LLM forecast. On Day~1, both execute the initial query policy.

\rbphase{Executed revisions}
The revisable plan in R1 removes broad \texttt{condition\_stats} acquisition on Days~2--4. On Day~5 it queries only the first listed runner in each race, then removes those queries again on Days~6--10. One saved intermediate version is replaced before formal execution.
\begin{lstlisting}[style=rbcasecode]
condition_stats query coverage:
    day 1:      all runners
    days 2-4:   none
    day 5:      first listed runner only
    days 6-10:  none
\end{lstlisting}

\rbphase{Recorded acquisition}
Frozen-$P_0$ makes 3,500 native condition-statistics acquisitions and purchases 1,136,800 cells through them. The revisable plan makes 287 and purchases 92,904 cells. Their weighted condition-statistics charges are 6,559.00 and 536.27 score-yen. Both arms make 4,568 depth-one recent-history acquisitions and 324 race-level workout purchases. Further local workout projections use the owned cache.

\textbf{Results (frozen $\rightarrow$ revisable evidence plan).}
Gross score: $14{,}502.13 \rightarrow 13{,}853.16$.
Weighted fees: $12{,}609.19 \rightarrow 6{,}586.46$.
Net score: $1{,}892.94 \rightarrow 7{,}266.70$.
All scores are total score-yen over 324 races. The arms generate their forecasts separately.
\end{rbcase}

\endgroup

\section{Reproducibility and Released Materials}
\label{app:repro}
The code repository is available at \url{https://github.com/HKUST-KnowComp/JockeyBench}.

Because the underlying JRDB archive is distributed under a commercial membership, raw database records and private HMAC entity mappings remain on the evaluation host.

\end{document}